\documentclass[3p,preprint,12pt]{elsarticle}

\usepackage{amssymb}

\usepackage{amsmath,lineno,hyperref,graphicx,booktabs,xcolor,multirow,textcomp}
\usepackage{cleveref}
\crefname{equation}{equation}{equations}
\Crefname{equation}{Equation}{Equations}
\crefrangelabelformat{equation}{(#3#1#4--#5#2#6)}
\newcommand\sg[1]{{\scriptsize\textcolor{gray}{$\pm$#1}}}
\defcitealias{donlapark20}{Ponnoprat, 2020}
\journal{Applied Soft Computing}

\begin{document}
\aboverulesep=0ex
\belowrulesep=0ex
\begin{frontmatter}



    \title{Short-term Daily Precipitation Forecasting with Seasonally-Integrated Autoencoder\footnote{\textcopyright \   2021. This manuscript version is made available under the CC-BY-NC-ND 4.0 license \\ \url{http://creativecommons.org/licenses/by-nc-nd/4.0/}}}


\author[address1]{Donlapark Ponnoprat\corref{cor1}}
\ead{donlapark.p@cmu.ac.th}
\cortext[cor1]{Corresponding author}
\address[address1]{Data Science Research Center, Department of Statistics, Faculty of Science, Chiang Mai University, Chiang Mai 50200, Thailand}

\begin{abstract}
    Short-term precipitation forecasting is essential for planning of human activities in multiple scales, ranging from individuals' planning, urban management to flood prevention. Yet the short-term atmospheric dynamics are highly nonlinear that it cannot be easily captured with classical time series models. On the other hand, deep learning models are good at learning nonlinear interactions, but they are not designed to deal with the seasonality in time series. In this study, we aim to develop a forecasting model that can both handle the nonlinearities and detect the seasonality hidden within the daily precipitation data. To this end, we propose a seasonally-integrated autoencoder (SSAE) consisting of two long short-term memory (LSTM) autoencoders: one for learning short-term dynamics, and the other for learning the seasonality in the time series. Our experimental results show that not only does the SSAE outperform various time series models regardless of the climate type, but it also has low output variance compared to other deep learning models. The results also show that the seasonal component of the SSAE helped improve the correlation between the forecast and the actual values from $4\%$ at horizon 1 to $37\%$ at horizon 3.
\end{abstract}




\end{frontmatter}


\section{Introduction}
\label{sec:intro}
Precipitation is one of the most essential weather phenomenon that impacts human activities in all scales, be it everyday commutes, urban constructions \citep{app01,app05}, flood control \citep{app07,app04,app06}, crop irrigation \citep{app03,app02} and hydroelectric power management \citep{app10,app08,app11,app09,app12}. Therefore, knowing the information about upcoming precipitation, especially in small timescale, is vital in successful planning. However, precipitation forecasts at hourly and daily timescales are considerably harder than those at monthly and seasonal scales due to high fluctuations of precipitation in a short period of time, especially during a wet season. 

In this study, we focus on a short-term forecast over the period of 1--3 days. In this range, one of the most common methods is the Numerical Weather Prediction (NWP), whose model consists of a system of differential equations that govern the evolution of weather systems through hydrodynamics and thermodynamics. Using the current weather state as an initial input, a forecast up to a few hours ahead can be obtained by solving these equations simultaneously. With recent advancement in computational power, the forecast can be made as often as every one hour with reasonable accuracy up to 12--hour lead time \citep{spin03,spin01}. Despite the advantages that make NWP a forecasting method trusted by government agencies and industries, it is difficult for general users to fully utilize the core model due to high computational cost and limited access to high-quality radar data. In extreme cases, the runtime of the model might even be longer than the forecasting period, creating the so-called spin-up problem \citep{dp08,dp07}. 

For those reasons, many efforts have been put into developing statistical methods that could extract the relationship between atmospheric variables from decades of past weather data. Since the actual forecasts from statistical methods do not involve any complicated differential equation solvers, they have an advantage of having low computational requirements and in some cases can perform better than NWP at very short forecasting periods. In terms of statistical forecasting at different timescales, classical time series models such as Autoregressive Integrated Moving Average (ARIMA) performs really well on series with clear seasonality and trend e.g. monthly and yearly data \citep{arima02,arima05,arima06,arima03,arima01,arima04}, but falls short for short-term forecasts due to high fluctuations and nonlinear nature of the data. For these reasons, some attention has turned to more complex models for time series forecasting, such as support vector regression \citep{Yu2006,Novitasari2020,Kisi2011,Joanna2020,Valente2020,Karasu2020}, random forest regression \citep{Fang2020,Peng2020,Callens2020,Ali2020,Karasu2019}, and artificial neural network. There is a series of works that studied neural networks and gave experimental results showing that they generally outperform linear models in very short- and short-term forecasts \citep{ann01,ann02,ann05,ann03,dp07,ann04,ann06}. 

With high complexity and well-refined optimization algorithms, these models have become methods of choice to solve regression problems. Nonetheless, their limitation lies in the lack of ability to express temporal relationships between data points, which makes them unsuitable for learning time series. On the other hand, the recurrent neural network (RNN) was specifically designed to handle sequential data by only allowing the information to flow forward in time. Thereafter, the long short-term memory (LSTM), proposed by \cite{dp02} in order to solve the instability problem of RNNs, has shown to be successful in various complex sequential tasks \citep{lstm01,lstm02,lstm03}. It is then not surprising that the LSTM would gain a lot of attention as a method for time series modeling \citep{Namini2018,Song2020,Masum2018,Zheng2017,Sahoo2019,Janardhanan2017,Altan2019}. For precipitation forecasting, many works have applied LSTMs to local meteorological data \citep{Poornima2019,Zhang2019,Aswin2018,Ni2020,Kumar2019,Kang2020} and high-dimensional data such as those from radar or satellite \citep{dp08,lstm04,1711.02316,Kumar2020}. On a related note, there have also been recent developments in other RNN-inspired time series models, such as Echo State Network \citep{Jaeger78,Yen2019,Ouyang2017}, Temporal Convolutional Network \citep{dp04} and Long- and Short-term Time-series Network \citep{dp03}. 

As it was originally designed, a single LSTM unit can only handle one-step ahead forecasts. To remove this limitation, \cite{lae01,lae02} proposed the LSTM autoencoder, which consists of two LSTM units: one for encoding the input sequences into a single vector, and one for decoding the vector into multiple forecasting values. This model has shown some success, for example, in hourly weather forecast \citep{s2s01} rainfall-runoff modeling \citep{Xiang2020} and flood forecasting \citep{Kao2020}.

The aim of this study is to develop an LSTM autoencoder-based model for daily precipitation forecasting. Note that at the daily scale, precipitation tends to have high fluctuation, while the seasonality is not clearly displayed; if we were to employ a single LSTM autoencoder for this task, it would prioritize learning the nonlinear dynamics and ignoring the seasonality hidden within the data. Hence, inspired by the time series methods with multiplicative seasonality, we propose a forecasting model, \emph{seasonally-integrated autoencoder} (SSAE), which consists of two LSTM autoencoders: one for learning short-term dynamics and one for learning the seasonality. Thus, the data has to be sent through the model twice in order to utilize both the short-term and the seasonal components. In addition, we have modified the LSTM autoencoders inside the SSAE so that it is more suitable to our forecasting task; the details of these modifications are given in the next section. To illustrate the effectiveness of this model, we compare its performance against various classical and deep learning models on two daily weather datasets: the data from Chiang Mai International Airport, Chiang Mai, Thailand, where the climate is tropical with the pronounced dry season, and the data from Theodore Francis Green State Airport, Providence, RI, where the climate is humid subtropical with evenly distributed precipitation throughout the year. 

\section{Methods}
\label{sec:method}

\subsection{Problem setting}
The goal of this study is to create a model that takes a sequence of multivariate time series input and forecasts a sequence of univariate time series output.  The input features consist of daily meterological data typically recorded in weather stations such as temperature, humidity, pressure, precipitation, wind speed etc. and the output is the daily total precipitation. From the data, we obtain $m$-dimensional input sequences of past $T$ days $X_t=(x_{t+1},x_{t+2},\ldots,x_{t+T})\in\mathbb{R}^{T\times m}$, and target sequences of future $H$ days $Y_t = (y_{t+T+1},y_{t+T+2},\ldots,y_{t+T+H})\in \mathbb{R}^H$. Our goal is to seek a forecasting model $f$ that satisfies
    \[
        Y_t = f(X_t).
    \] 
Here, we are interested in short-term forecasts where $1\leq H \leq 3$.

\subsection{Long short-term memory (LSTM)}

\begin{figure}[t]
    \centering
    \includegraphics[width=0.8\columnwidth]{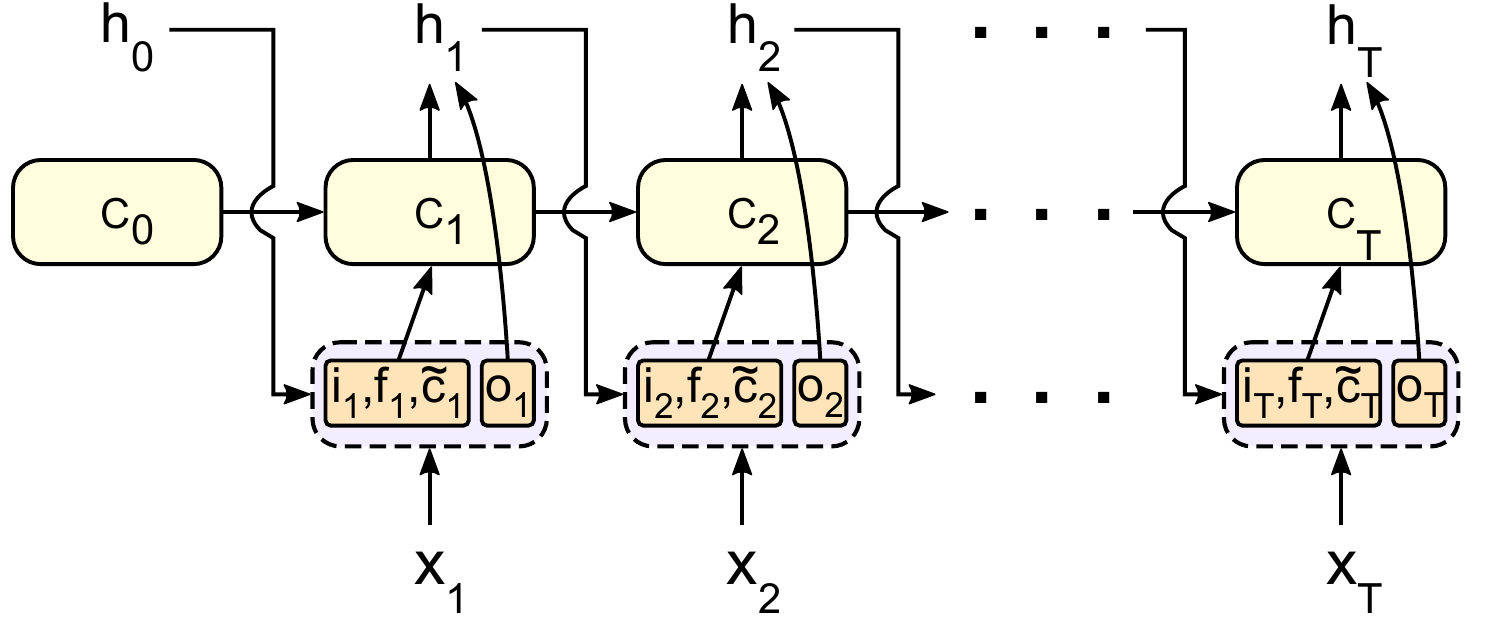}
    \caption{A single-layer LSTM which takes an input sequence $x_1,x_2,\ldots,x_T$ and return a sequence $h_1,h_2,\ldots,h_T$. The connection weights and activation functions are omitted.}
    \label{fig:lstm}
\end{figure}

LSTM was proposed in \cite{dp02} as a solution to gradient vanishing problem in recurrent neural networks. An LSTM cell has a \emph{cell state} $c_t$ that memorizes information from the past, which is then managed by the following gate units:
\begin{itemize}
    \item forget gate unit $f_t$ which carries some of the past information from $c_{t-1}$
    \item input gate unit $i_t$ which contains the amount of new information $\tilde{c}_t$
    \item output gate unit $o_t$ which activates the new cell state $c_t$ in order to obtain the hidden state $h_t$.
\end{itemize}

Formally, if the number of hidden units is $h$, then all of these units are elements of $\mathbb{R}^h$. The following computations take place in each time step:
\begin{subequations}\label{eq:subs}
    \begin{align}
    f_t &= \sigma(U_fx_t+W_fh_{t-1}+b_f) \label{eq:sub1} \\
    i_t &= \sigma(U_ix_t+W_ih_{t-1}+b_i)\label{eq:sub2} \\
    o_t &= \sigma(U_ox_t+W_oh_{t-1}+b_o)\label{eq:sub3} \\
    \tilde{c}_t &= \phi(U_cx_t+W_ch_{t-1}+b_c) \label{eq:sub4}\\
    c_t &= f_t\odot c_{t-1} + i_t\odot \tilde{c}_t \label{eq:sub5}\\
    h_t &= o_t\odot \phi(c_t), \label{eq:sub6}
\end{align}
\end{subequations}
where $\odot$ is the elementwise vector product, $\sigma$ is the sigmoid function, $\phi$ is an activation function, $U\in\mathbb{R}^{h\times m}$, $W\in\mathbb{R}^{h\times h}$ and $b\in\mathbb{R}^h$. In our proposed model, we chose $\phi$ to be the Rectifier Linear Unit (ReLU):
\[
    \phi(x)=\text{ReLU}(x) = \max(0,x).
\] 
The causal relationships between the units in LSTM are illustrated in \Cref{fig:lstm}.

\subsection{LSTM autoencoder}
\begin{figure}[t]
    \centering
    \includegraphics[width=0.85\columnwidth]{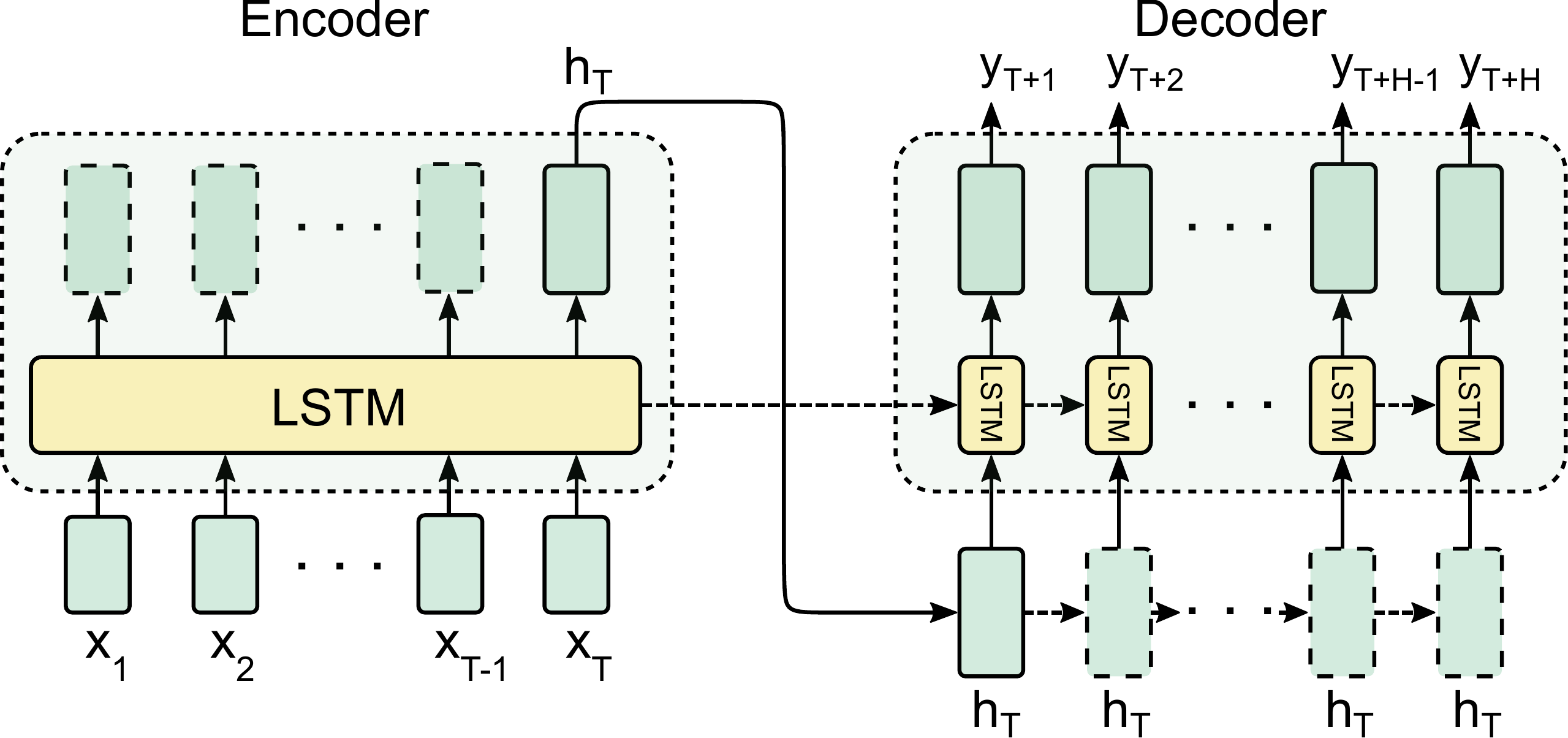}
    \caption{A modified LSTM autoencoder (S2S--2) used in our proposed model.}
    \label{fig:encoder}
\end{figure}
The LSTM autoencoder \citep{lae01,lae02,s2s01} has two LSTM blocks; the first block is an \emph{encoder} which turns an input sequence into a single vector, which can be considered as a hidden representation of the input data, The second block is a decoder that turns the vector into an actual prediction (\Cref{fig:encoder}). As there are multiple units in the decoder and only one hidden vector, there are multiple ways to design how information flows in the decoder. For example, each LSTM unit in the decoder can take a copy of the hidden vector, or it can takes the output of the previous unit as the input. In our proposed model, we give the hidden vector to each of the LSTM units in the decoder. Additionally, each time step in the decoder has its own LSTM unit, meaning that the weights $U$ and $W$ in \cref{eq:sub1,eq:sub2,eq:sub3,eq:sub4} are not shared across these units. We call this modification of the LSTM autoencoder \emph{sequence-to-sequence 2} (S2S--2).

\subsection{Seasonally-integrated autoencoder (SSAE)}
\begin{figure}[t]
    \centering
    \includegraphics[width=\columnwidth]{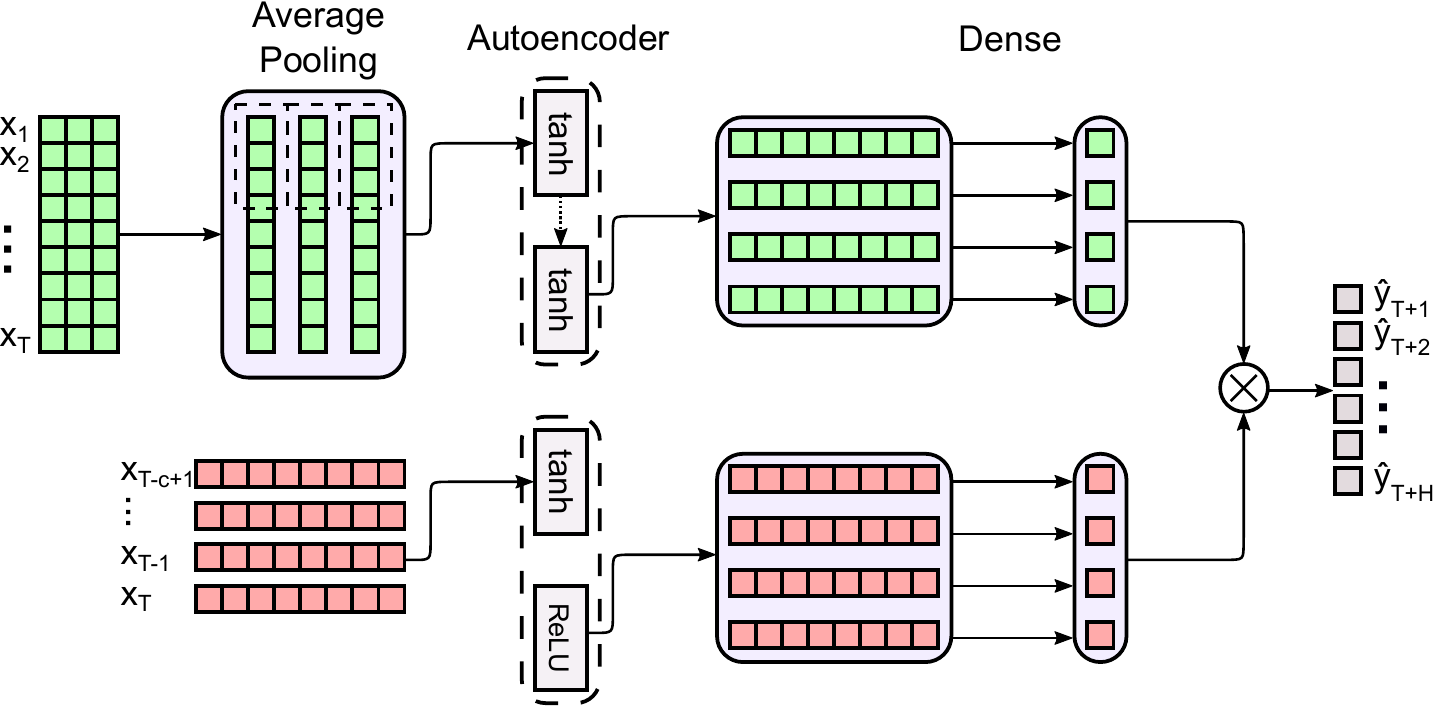}
    \caption{The proposed model for short-term precipitation forecasting. Note the difference in the activation functions of the autoencoders.}
    \label{fig:main_model}
\end{figure}

For precipitation forecasting, we propose \emph{seasonally-integrated autoencoder} (SSAE). The model consists of two multiplicative components: the first component makes a forecast based on short-term dynamics between the variables, while the second component detects seasonal effects that appear in the time series. Each of these components is an LSTM autoencoder with the following configurations (\Cref{fig:main_model}):
\begin{itemize}
    \item Short-term autoencoder $f_S$, which takes the last $c$ entries of the original training sequence $X^c_t=(x_{t-c+1+T},\ldots,x_{t+T})$ as its input. It has the following internal structure: 
\begin{itemize}
    \item The final hidden states of the encoder are not fed to the decoder.
    \item The encoder has $\tanh$ as its activation function, while the decoder has $\text{ReLU}$.
\end{itemize}
    \item Seasonal autoencoder $f_L$, which takes the full sequence of length $T$ as its input and has no activation function. In contrast to the short-term counterpart, $f_L$ has the following internal structure:
\begin{itemize}
    \item The final hidden states of the encoder are fed to the decoder.
    \item Both the encoder and the decoder have $\tanh$ as their activation function.
\end{itemize}
\end{itemize}
Since it is redundant to use all features to learn seasonal patterns, only a few of them will be selected to train $f_L$. We also need to remove any short-term fluctuations by smoothing so that the model has an easier time learning global patterns. Let $z_t$ be the data of said features at time $t$. Then, the input of the seasonal autoencoder can be written as $Z_t=(z_t,z_{t+1},\ldots,z_{t+T})$. Before applying the autoencoder, we compute the moving average of each sequence in the training data:
\[ \bar{z}_t =\frac{z_t+z_{t-1}+\ldots+z_{t-\ell+1}}{\ell}=\frac{1}{\ell}\sum_{i=0}^{\ell-1}z_{t-i}, \]
where $\ell$ is the moving window. We also set the stride value $\Delta$, that is, the step size between windows to reduce the dimension of input data, and so the computation time. Conveniently, this technique is available in many deep learning libraries such as Tensorflow and Pytorch as an \emph{average pooling} layer, which can be placed directly as a part of the network so that we can take advantage of well-optimized computation. After this layer, the transformed data $\overline{Z}_t^{\Delta}=(\ldots,\bar{z}_{t-2\Delta},\bar{z}_{t-\Delta},\bar{z}_t)$ then becomes an input to the seasonal autoencoder $f_L$. 

At the end of each LSTM autoencoders, we have a sequence of $H$ hidden vectors, which are then linearly transformed into $H$ scalars with a dense layer. Let $W_S$ and $W_L$ be the linear transformations that represent the dense layer placed after $f_S$ and $f_L$, respectively, then the short-term and seasonal outputs are given by
\begin{align*}
    g_S(X_t) &= W_Sf_S(X^c_t) \\
    g_L(X_t) &= W_Lf_L(\overline{Z}_t^{\Delta}), 
\end{align*}
and the precipitation forecast at horizon $H$ is  
\begin{equation} \widehat{Y}_t = g_S(X_t)\odot g_L(X_t),\label{mainmodel} \end{equation}
where $\odot$ is the element-wise matrix product.

\subsection*{Hyperparameter selection}

The SSAE has many hyperparameters that need to be adjusted. In general, suitable choices of hyperparameters depend on the characteristics of the time series, such as time-frequency, short-term fluctuations, and seasonality period. For the daily precipitation forecasting, our experimental results and a good rule of thumb \citep{s2s01,Reimers2017} suggest that the number of hidden units of all LSTMs should be around $100$. The input length of the short-term autoencoder should not be more than $7$ days; if the climate has no wet and dry seasons, we recommend to not use more than two or three days. The input length of the seasonal autoencoder should be the number of days in a wet or dry season. If that is not available, the input length should be between $60-120$ days. The average-pooling window ($\ell$)  should be between $1/3$ and $2/3$ of the input length, and the stride value ($\Delta$) should be somewhere between $10$ and a half of this window. To find the best set of hyperparameters among these combinations, we recommend the random search approach, optimized by the cross-validation.

\subsection*{Complexity analysis}

We now calculate the computational complexity of the SSAE. Let $d_S$ and $h_S$ be the input dimension and the hidden units of the short-term autoencoder, and $d_L$ and $h_L$ be those of seasonal autoencoder. From \eqref{eq:subs}, each LSTM unit requires $O(d_Sh_S+h_S^2)$. There are $c$ input units in the encoder and $H$ input units in the decoder. Note also that there is a dense layer $W_S\in\mathbb{R}^{H\times h_S}$ which contributes $O(h_SH)$ to the complexity. Thus the complexity of the autoencoder is $O\left( \left(d_Sh_S+h_S^2  \right)(c+H)  \right)$. Similarly, the complexity of the seasonal autoencoder is $O\left( \left(d_Lh_L+h_L^2  \right)\left(\frac{T}{\Delta}+H\right)  \right)$. Hence, the complexity of the SSAE per time sequence is $O\left( \left(d_Sh_S+h_S^2  \right)(c+H)+ \left(d_Lh_L+h_L^2  \right)\left(\frac{T}{\Delta}+H\right)  \right)$.

Next, we analyze the number of parameters. Recall that the weights are shared among the LSTMs inside the encoder but not the decoder. Thus most of the parameters are contained in the decoders. From the calculations above, the SSAE has $O\left( \left(d_Sh_S+h_S^2  + d_Lh_L+h_L^2  \right)H  \right)$ number of parameters. Hence, besides the input dimension and the number of hidden units, the training time is also heavily affected by the forecasting horizon.

\section{Data preprocessing}
The data cannot yet be fed directly into the model as there are issues of input format, scales of variables, and possible outliers. We address these issues by preprocessing data with the following techniques.

\subsection{Feature scaling}
There are two traditional choices when it comes to scaling time series: min-max scaling, where the features are scaled to fit between zero and one, and standardization, where they are scaled to have zero mean and unit variance. Looking back at the equation \eqref{mainmodel} of our proposed model, any small perturbation that changes the sign of an output of either the short-term or the seasonal component would flip the sign of the forecast, leading to an unstable training algorithm. Therefore, we chose min-max scaling; for the value $x_{jt}$ of the $j$-th feature at time $t$, we make the following transformation:
 \[
     x_{jt}^{\text{new}} = \frac{x_{jt}-x^{\text{min}}_{j}}{x^{\text{max}}_{j}-x_j^{\text{min}}},
\] 
where $x_j^{\text{min}}$ and $x_j^{\text{max}}$ are the minimum and the maximum value of the $j$-th feature, respectively.

\subsection{Moving window method}

To train the model, we need to transform the data into input sequences of lengths $T$ and target sequences of length $H$. This can be done by sequentially cropping the original time series $x_1,$$x_2,$$\ldots,$$x_N \in \mathbb{R}^{N\times m}$ and $y_{1},$$y_{2},$$\ldots,$$y_N\in\mathbb{R}$ with adjacent temporal windows of size $T$. For example, the first instance would be $(x_1,x_2,\ldots, x_T)$ and $(y_{T+1},y_{T+2}\ldots,y_{T+H})$. All training instances are obtained by moving both windows simultaneously by one unit until the target window reaches the end of the target series, resulting in $N-T-H+1$ input and target sequences. For convenience, we will refer to each of the input-target pairs by the first date of the target sequence.

\section{Minibatch training}
Let $N$ be the number of instances in the training set. In one round of training (one \emph{epoch}) one usually randomly splits the training set into smaller \emph{minibatches} of $M$ instances to reduce the memory requirement. Suppose that $\{X_{N_i}\}_{i=1}^M$ and $\{Y_{N_i}\}_{i=1}^M$ is a minibatch. We measure the error of the prediction $\widehat{Y}_{N_i}$ by a \emph{loss function} $l(\widehat{Y}_{N_i},Y_{N_i})$. A sequence-to-sequence deep learning model is trained by adjusting the connection weights so that the objective function
\begin{align}\label{lossfunc}
    L &= \frac{1}{M}\sum_{i=1}^M l(\widehat{Y}_{N_i},Y_{N_i}), \\
        \intertext{as a function of the weights, is minimized. For regression tasks, the most frequently used loss function is the mean squared error (MSE)} 
    l(\widehat{Y}_{N_i},Y_{N_i}) &= \frac{1}{H} \sum_{k=1}^H (\hat{y}_{N_i+T+k}-y_{N_i+T+k})^2.\label{eqn:mse}
\end{align} 
The minimization of $L$ is done via backpropagation \citep{bp}: at the first step, each weight $w$ is randomly initialized as $w_0$, and during $k$-th step, it is adjusted by a small fraction $\eta$ of the gradient of the loss function:
\begin{equation} w_{k} = w_{k-1}-\eta\frac{\partial L}{\partial w}(w_{k-1}).\label{update}  \end{equation}
 The small number $\eta$, typically called the \emph{learning rate}, dictates how far the weights move in each step on the curve of $L$. Hence, one must be careful when choosing a value of $\eta$. If it is too large, the weight will never reach a minimum; if it is too small, they will get stuck at a local but not global minimum.

\section{Experiments}
\label{sec:experiments}
In the experiments, we trained SSAE on two separate datasets and compare its performance against several benchmarks. The classical models include Vector Autoregression Moving-Average (VARMA), Support Vector Regression (SVR), and Gradient Boosting (GB). We also compared SSAE against recently proposed time series models:

\par{\textbf{Temporal Convolutional Network (TCN)} by \cite{dp04}}: A sequence-to-sequence model consisting of dilated causal convolutional layers \citep{Oord16} which only allows past values to propagate forward. It also utilizes residual connections which allow the network to have a large number of parameters without overfitting.
\par{\textbf{Long- and Short-term Time-series Network (LSTNet)} by \cite{dp03} }: A deep learning model that starts with convolutional layers to extract short-term dependency among variables, following by recurrent layers to discover seasonal patterns within the time series.
      \par{\textbf{Echo State Network (ESN)} by \cite{Jaeger78}}: A recurrent neural network with one hidden \emph{reservoir layer} which consists of sparsely connected nodes. The weights between the input and the reservoir are fixed and cannot be trained, while those between the reservoir and the output are trainable. Therefore, the training procedure is faster than the typical recurrent neural network with the same number of hidden nodes. The implementation of ESN that we used in the experiments was from \cite{echotorch}.

      \par{\textbf{Sequence-to-sequence 1 (S2S--1)}}: The original LSTM autoencoder that was used in \cite{s2s01} for sequence-to-sequence hourly weather forecast. Note that this model also has an extra $100\rightarrow 100$ dense layer after the encoder.

      \par{\textbf{Sequence-to-sequence 2 (S2S--2)}}: Our modified version of the LSTM autoencoder. Each LSTM unit in the decoder has its own weights, and the forecast from $i$--th unit is fed as an input to the $i+1$--th unit. In addition, the $\tanh$ activation function is replaced by the ReLU function. This is the type of autoencoder that we use in the SSAE.

        One advantage of autoencoder models (S2S--1, S2S--2 and, SSAE) is that their designs allow the forecasting horizon to be extended naturally. However, this is not the case for any of the remaining models besides VARMA. For these models, we had to apply the following sequential procedures: if $f$ is the model for day-1 forecast and $x_{t+1},x_{t+2},\ldots,x_{t+T}\in \mathbb{R}^m$ is an input sequence, then the in-sample forecast is
        \[
            \hat{x}_{t+T+1} = f(x_{t+1},x_{t+2},\ldots,x_{t+T}).
        \] 
        To make a day-2 forecast, we use $\hat{x}_{t+T+1}$ instead of the actual value $x_{t+T+1}$. Hence, the next prediction is
        \[  \hat{x}_{t+T+2} = f(x_{t+2},\ldots,x_{t+T},\hat{x}_{t+T+1}),           \]
    and so on. 

    In each experiment, the last three years of the data was set apart as a held-out test set for model evaluation. The remaining data was further split into a training set and a validation set for hyperparameter tuning via random searches. 

Let $\tilde{N}$ be the number of instances in the test set. Denote the actual sequence of future precipitation by
\begin{align*}Y_t&=(y_{t+T+1},\ldots,y_{t+T+H}), \hspace{20px} t=1,2,\ldots,\tilde{N}, \\
    \intertext{and the model's predictions by}
    \widehat{Y}_t &= (\hat{y}_{t+T+1},\ldots,\hat{y}_{t+T+H}), \hspace{20px} t=1,2,\ldots,\tilde{N}. 
\end{align*}
    We evaluated the performance of the model at each forecasting horizon $H$ using the following two metrics: 
    \begin{align*}
    \intertext{Root-mean-squared error (RMSE):}
\text{RMSE} &= \sqrt{\sum_{t=1}^{\widetilde{N}} (y_{t+T+H}-\hat{y}_{t+T+H})^2}.  
        \intertext{Pearson's correlation coefficient (CORR):}
    \text{CORR} &= \frac{\sum_{t=1}^{\widetilde{N}}(y_{t+T+H}-\bar{y}_{T+H})(\hat{y}_{t+T+H}-\bar{\hat{y}}_{T+H})}{\sqrt{\sum_{t=1}^{\widetilde{N}}(y_{t+T+H}-\bar{y}_{T+H})^2}\sqrt{\sum_{t=1}^{\widetilde{N}}(\hat{y}_{t+T+H}-\bar{\hat{y}}_{T+H})^2}},
\end{align*}
where $\bar{y}_{T+H}$ is the average of $\{y_{t+T+H}\}_{t=1}^{\tilde{N}}$ and $\bar{\hat{y}}_{T+H}$ is the average of $\{\hat{y}_{t+T+H}\}_{t=1}^{\tilde{N}}$.

We evaluated the models on two daily weather datasets in order to see how well they perform on different types of climate. For consistent results from deep learning models, several learning techniques have been used to reduce the variance of the predictions; see \ref{sec:trainalg}. Across all deep learning models, the batch size (the number of samples per one training iteration) was set to $256$.

The datasets, along with the python source code of three autoencoder models, S2S--1, S2S--2, and SSAE, are hosted at \citepalias{donlapark20}.

\subsection{Precipitation at Chiang Mai International Airport, Thailand}

The first dataset was provided by the Thai Meteorological Department\footnote{\url{https://www.tmd.go.th/en}} via personal communication. The data had been collected at Chiang Mai International Airport from January 1st, 1998 to July 31st, 2019. All redundant variables were removed, and the remaining variables were selected to train the models (\Cref{table:cmfeat}).
\begin{table}[!htp]
    \centering\footnotesize
    \begin{tabular}{ll}
        \toprule
        Variable& Value range \\
        \hline 
        Daily minimum pressure & 904--1022 hPa \\
        Daily maximum pressure & 1000--1028.12 hPa\\
        Daily minimum temperature & 3.8--29.3 $^{\circ}$C\\
        Daily maximum temperature & 12.1--29.3 $^{\circ}$C\\
        Daily minimum relative humidity & 9--98\%\\
        Daily maximum relative humidity & 44--99\%\\
        Daily total evaporation &0--36.2 mm\\
        Daily sunshine duration &0--12.8 hours\\
        Daily average wind speed & 0--35.5 m/s\\
        Daily average wind direction & 10--360$^{\circ}$\\
        Daily total precipitation &0--144 mm\\
        \bottomrule
\end{tabular}
    \caption{Meteorological variables of the data at Chiang Mai International Airport and their value ranges.}
    \label{table:cmfeat}
\end{table}

The training set consists of the data from January 1st, 1998 to July 31st, 2016, and the test set consists of the data from August 1st, 2016 to July 31st, 2019. The VARMA, SVR, and GB were trained only once, since their training algorithms are deterministic. The deep learning models, on the other hand, are stochastic due to the random weight initialization and random batch sampling. To account for variance in test scores, we trained each of these models 30 times and calculated the scores' statistics. 

For SSAE, the daily minimum and maximum relative humidity were chosen to model the seasonal effect. The seasonal look-back window ($T$) was $169$ days. The short-term window ($c$) was $6$ days. The number of hidden units was $100$ for both seasonal and short-term components of the model. The average moving window ($\ell$) was $125$ days, and the stride value ($\Delta$) was $60$ days. The hyperparameters of the other models are provided in \ref{sec:hyper}.

\begin{table}[!htp]
    \centering\small
    \begin{tabular}{@{}lllllllll@{}}
       \toprule 
        & \multicolumn{2}{c}{Day 1}                           & \multicolumn{1}{c}{} & \multicolumn{2}{c}{Day 2}                           & \multicolumn{1}{c}{} & \multicolumn{2}{c}{Day 3}                           \\ \cline{2-3} \cline{5-6} \cline{8-9} 
       Model   & \multicolumn{1}{c}{RMSE} & \multicolumn{1}{c}{CORR} & \multicolumn{1}{c}{} & \multicolumn{1}{c}{RMSE} & \multicolumn{1}{c}{CORR} & \multicolumn{1}{c}{} & \multicolumn{1}{c}{RMSE} & \multicolumn{1}{c}{CORR} \\ \midrule
VARMA     & 7.04 & 0.398 &    &  7.34 & 0.286 &  & 7.37 & 0.272  \\
SVR     & 8.29 & 0.213 &   & 8.50 & 0.147 &  & 8.47  & 0.142   \\
GB     & 7.08 & 0.382 &   & 7.42 & 0.285 &  & 7.51 & 0.270    \\
TCN     &    7.82\sg{0.37} & 0.170\sg{0.052} & & 7.55\sg{0.10} & 0.175\sg{0.076} & & 7.58\sg{0.19} & 0.181\sg{0.100}          \\
LSTNet     & 7.10\sg{0.05} & 0.380\sg{0.011} & & 7.42\sg{0.06} & 0.271\sg{0.011} & & 7.43\sg{0.05} & 0.264\sg{0.013}      \\
ESN     & 7.15\sg{0.05} & 0.363\sg{0.013} & & 7.43\sg{0.08} & 0.271\sg{0.024} & & 7.55\sg{0.19} & 0.251\sg{0.033}    \\
S2S--1  & 7.03\sg{0.02} & 0.395\sg{0.006} & & 7.33\sg{0.02} & 0.297\sg{0.006} & & 7.44\sg{0.04} & 0.260\sg{0.012}     \\
S2S--2  & 6.98\sg{0.01} & 0.410\sg{0.004} & & 7.30\sg{0.01} & 0.305\sg{0.003} & & 7.38\sg{0.01} & 0.276\sg{0.005}    \\
\textbf{SSAE} &  \textbf{6.96}\sg{0.01} & \textbf{0.418}\sg{0.002} & & \textbf{7.28}\sg{0.01} & \textbf{0.311}\sg{0.003} & & \textbf{7.35}\sg{0.01} & \textbf{0.287}\sg{0.006}     \\ 
\bottomrule
\end{tabular}%
\caption{The RMSE and CORR scores of day 1--3 forecasts. For all deep learning models, the mean and sample standard deviation of the scores from 30 training sessions are reported.}
\label{table:cmresults}
\end{table}

\begin{figure}[!htp]
    \centering
    \includegraphics[width=\columnwidth]{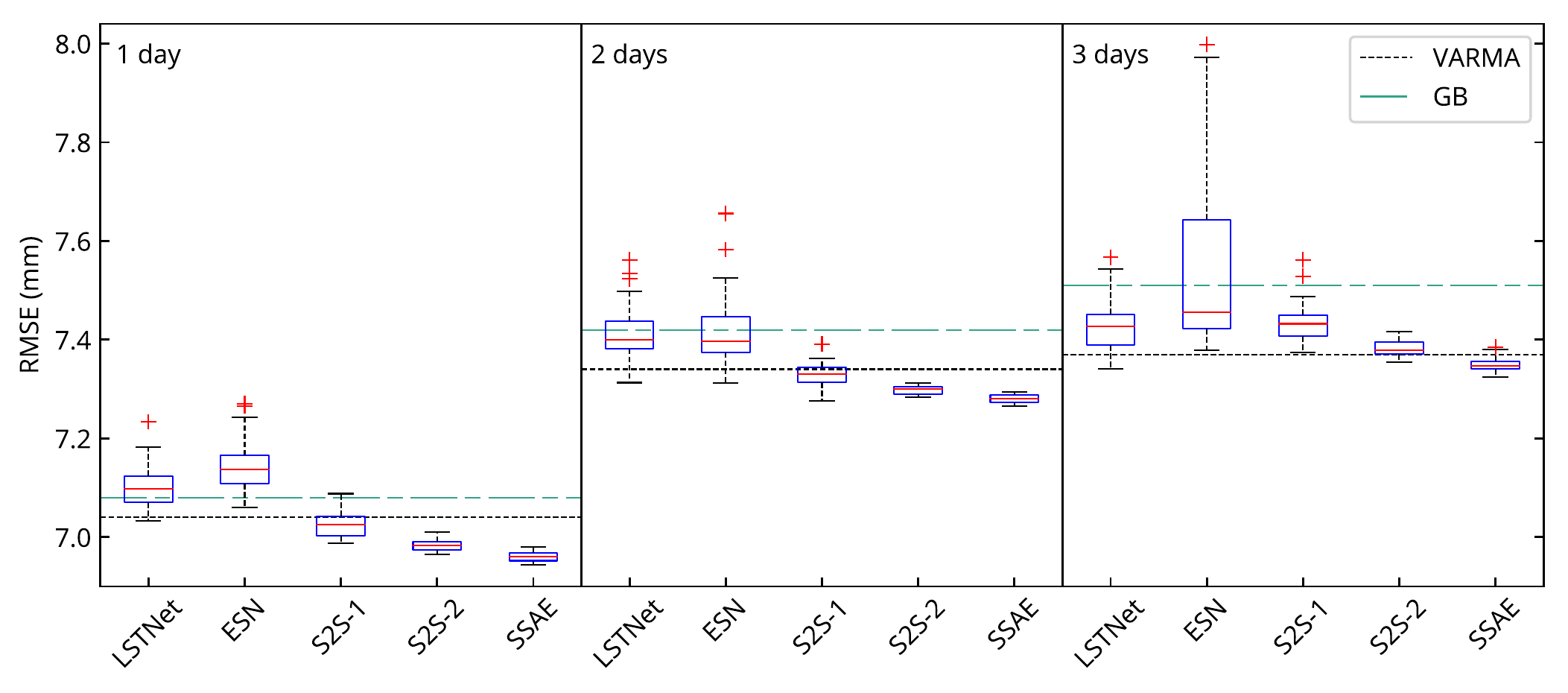}
    \includegraphics[width=\columnwidth]{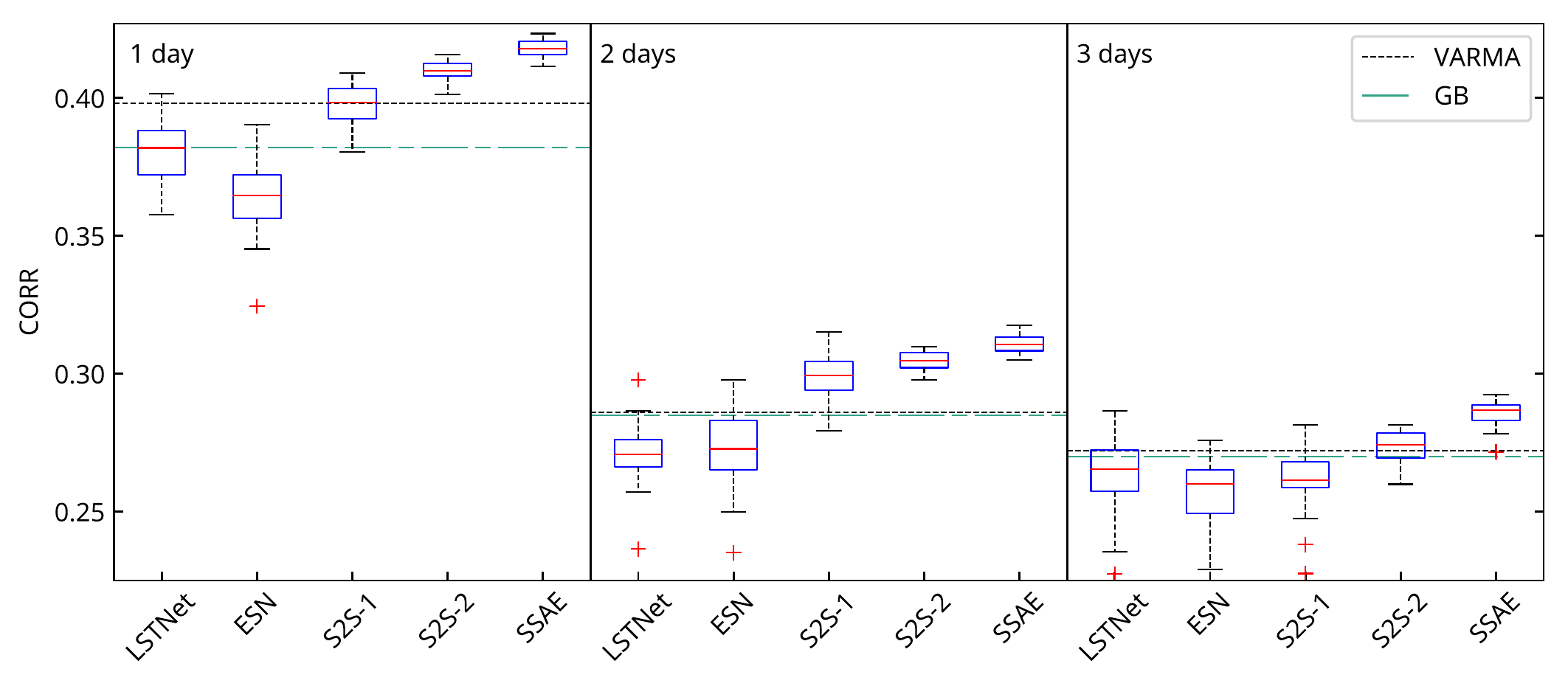}
    \caption{Performance of seven forecasting models on the test set. The scores of the SVR and TCN are not shown as they performed significantly worse than the other models.}
    \label{fig:cmbox}
\end{figure}

The RMSE and CORR of the forecasts on the test set, shown in \Cref{table:cmresults} and \Cref{fig:cmbox}, indicates that the SSAE outperforms other models on this task. Note that the S2S--2 is just the short-term component of the SSAE. By comparing these two models, we see that the seasonal component of the SSAE helped improve the RMSE by $0.3\%-0.6\%$ and the CORR by $2\%-4\%$. Moreover, we also notice from \Cref{fig:cmbox} that the output variances of the SSAE and S2S--2 are smaller than those of the other deep learning models. 

\begin{figure}[!htp]
    \centering
    \includegraphics[width=\columnwidth]{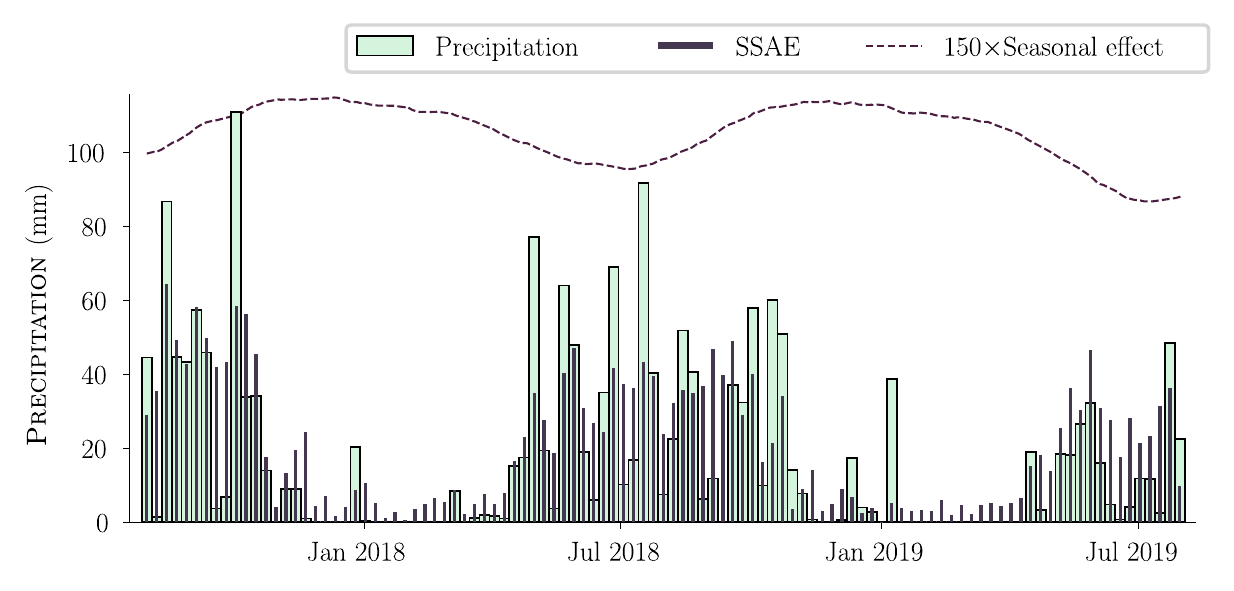}
    \caption{Weekly aggregated precipitation (mm), SSAE's next-day forecasts and the outputs of seasonal autoencoder from August 1st 2017 to July 31st 2019.}
    \label{fig:cmpred}
\end{figure}

\subsubsection*{Variable importance}

\begin{figure}[!t]
    \centering
    \includegraphics[width=\columnwidth]{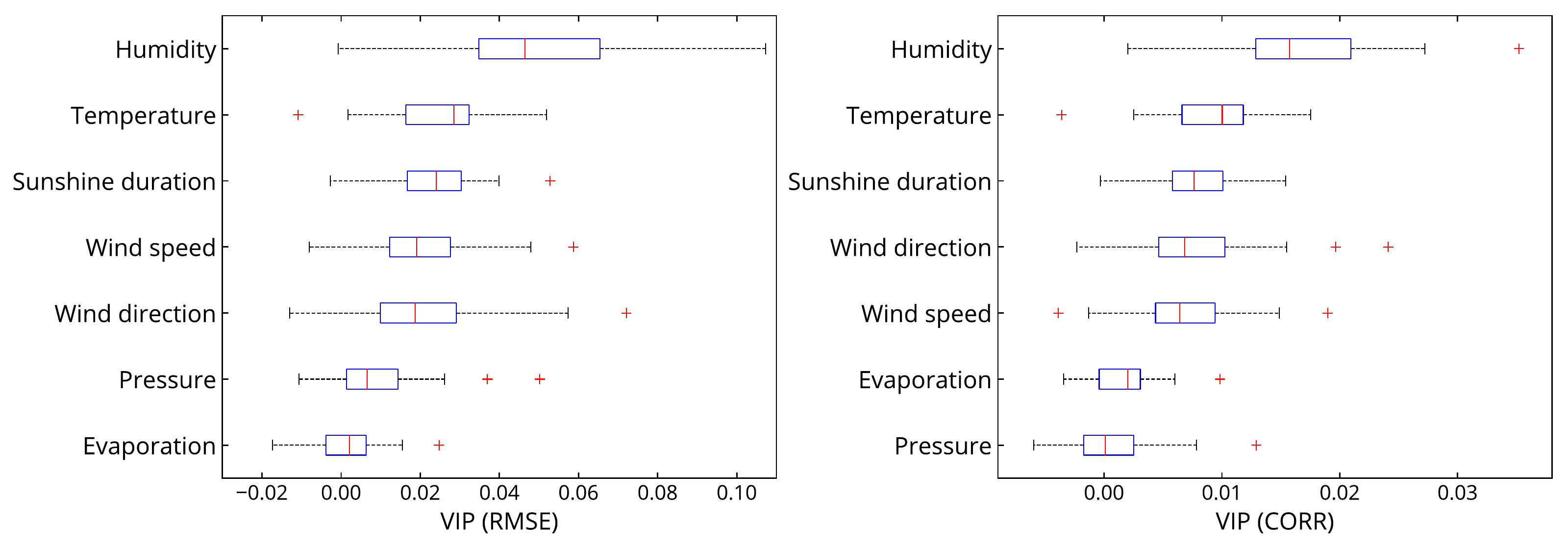}
    \caption{The variable importance scores measured by the RMSE and CORR scores of next-day forecasts on the test set.}
    \label{fig:weather_vip}
\end{figure}

We now measure SSAE's dependency on each variable by adopting the \emph{permutation feature importance}. In this method, we shuffle one of the variables, train the model, then obtain the forecasting scores. The variable importance score (VIP) is measured by the decrease in the scores from the mean values in \Cref{table:cmresults}. In this case, we use next-day RMSE and CORR to compute the variable importance score; let $R_j$ and $C_j$ be the new RMSE and CORR obtained after shuffling the variable $X_j$. Then the importance score of $X_j$ with respect to RMSE is  
\begin{align*}
    \text{VIP}^{\text{R}}_j &= R_j-6.96,
    \intertext{and the importance score of $X_j$ with respect to CORR is}
    \text{VIP}^{\text{C}}_j &= 0.418-C_j.
\end{align*}
The results of 30 repeated training sessions are shown in \Cref{fig:weather_vip}. We can see that under this criterion, the humidity is the most important variable.

\subsubsection*{Predictive uncertainty}

\begin{figure}[t]
    \centering
    \includegraphics[width=0.9\columnwidth]{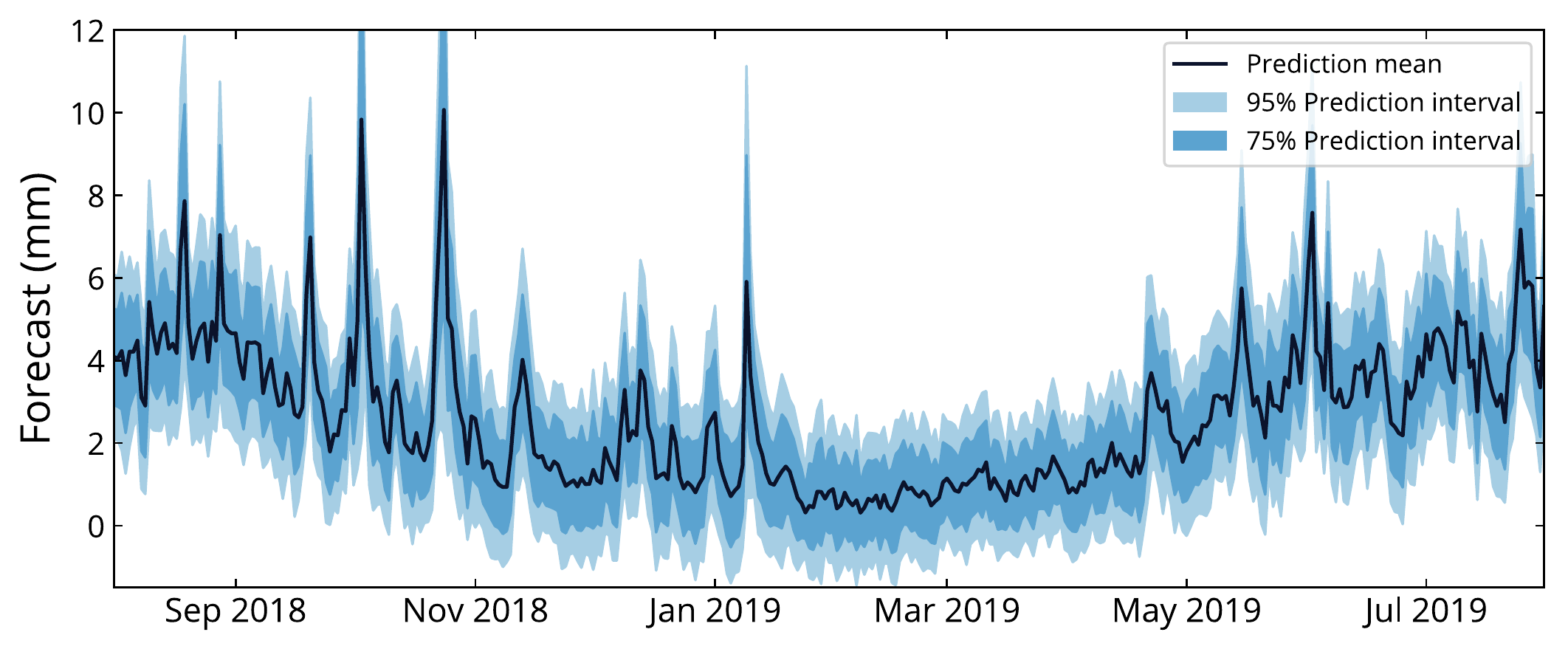}
    \caption{The confidence intervals of the next-day forecasts obtained via Monte Carlo dropout.}
    \label{fig:weather_conf}
\end{figure}

There is always uncertainty in a neural network's predictions due to the random initialization of the weights. In the context of precipitation forecasting, the confidence of the prediction is crucial for planning, especially for risk-averse planners. To determine the predictive uncertainty, we employ the \emph{Monte Carlo dropout}; during each iteration of training and prediction, each hidden node in an SSAE has a probability $p$ to be dropped out from the network. It can be shown that the resulting model is an approximation of a Gaussian process \citep{Gal2016}. Thus, by performing repeated predictions on the same data point, we can infer an estimated predictive confidence interval of that point. In this experiment, we ran SSAE 50 times on the test set with $p=0.25$.The $75\%$ and $95\%$ confidence intervals are shown in \Cref{fig:weather_conf}.

\subsection{Precipitation at Theodore Francis Green State Airport, Providence, RI}

The second dataset was acquired via online requests from the website of the National Oceanic and Atmospheric Administration\footnote{\url{https://www.ncdc.noaa.gov/cdo-web/datatools/lcd}}. The data had been collected at the Theodore Francis Green State Airport, Providence, RI from January 1st, 2006 to October 31st, 2019. All redundant variables were removed, and the remaining variables were selected to train the models (\Cref{table:pvfeat}).

\begin{table}[!htp]
    \centering\footnotesize
    \begin{tabular}{ll}
        \toprule
        Variable& Value range \\
        \hline 
        Daily average pressure & 980--1041 hPa\\
        Daily minimum temperature & -22.78--25.56 $^{\circ}$C \\
        Daily maximum temperature & -12.22--38.89 $^{\circ}$C \\
        Daily average dew point temperature & -27.78--25 $^{\circ}$C \\
        Daily average relative humidity & 19--100\% \\
        Daily average wind speed & 0.13--13.86 m/s\\
        Daily peak wind speed &2.68--29.50 m/s\\
        Daily peak wind direction &10--360$^{\circ}$\\
        Daily sustained wind speed & 2.24--21.46 m/s \\
        Daily sustained wind direction & 10--360$^{\circ}$\\
        Daily total precipitation &0--135.13 mm\\
        \bottomrule
\end{tabular}%
    \caption{Meteorological variables of the data at Theodore Francis Green State Airport and their value ranges.}
    \label{table:pvfeat}
\end{table}

In this experiment, the training set consists of the data from January 1st, 2006 to October 31st, 2016, and the test set consists of the data from November 1st, 2016 to October 31st, 2019. As in the previous experiment, each of the deep learning models was trained 30 times. 

For SSAE, the daily peak and sustained wind direction were chosen to model the seasonal component. The seasonal look-back window was $70$ days. The short-term window was $2$ days. The number of hidden units was $120$ in the seasonal autoencoder and $100$ in the short-term autoencoder. The average moving window was $41$ days, and the stride value was $14$ days. See \ref{sec:hyper} for the hyperparameters of the other models. 

\begin{figure}[!htp]
    \centering
    \includegraphics[width=\columnwidth]{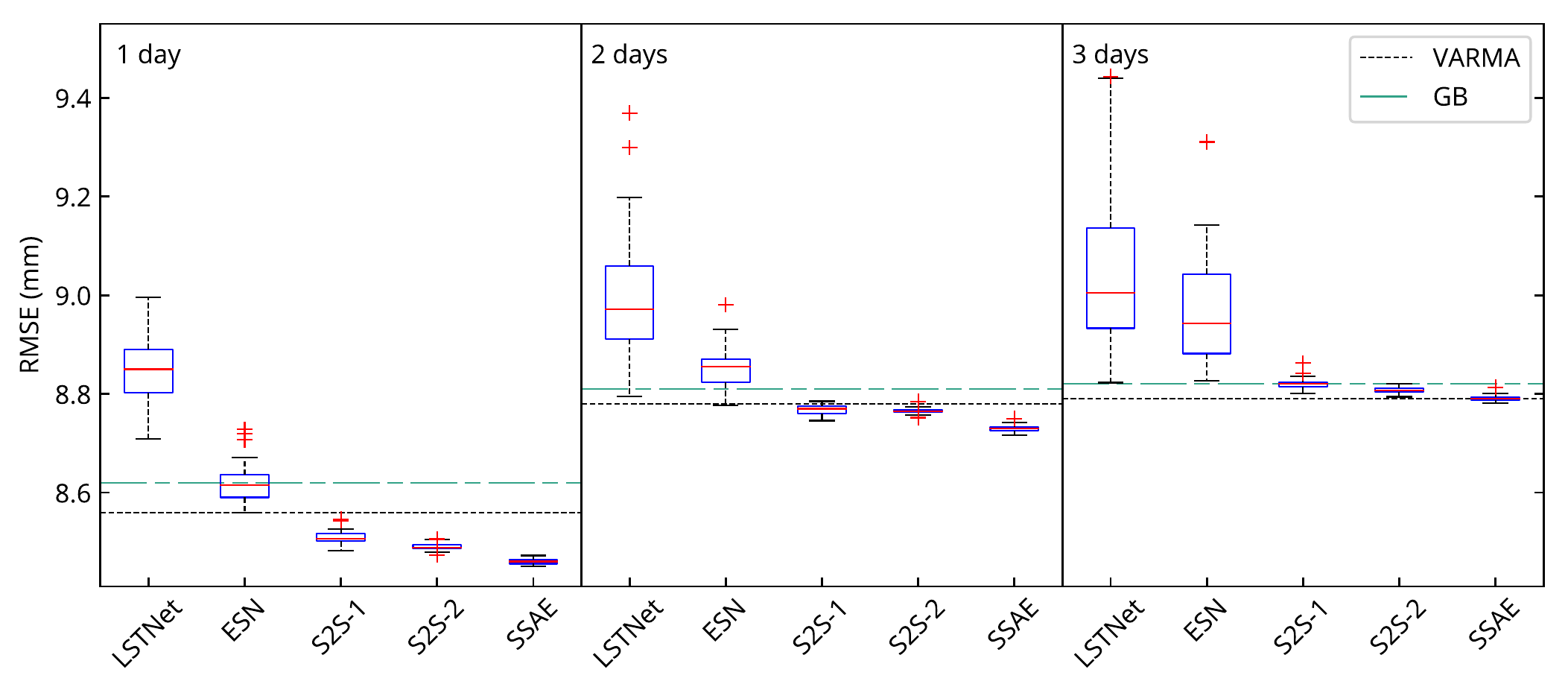}
    \includegraphics[width=\columnwidth]{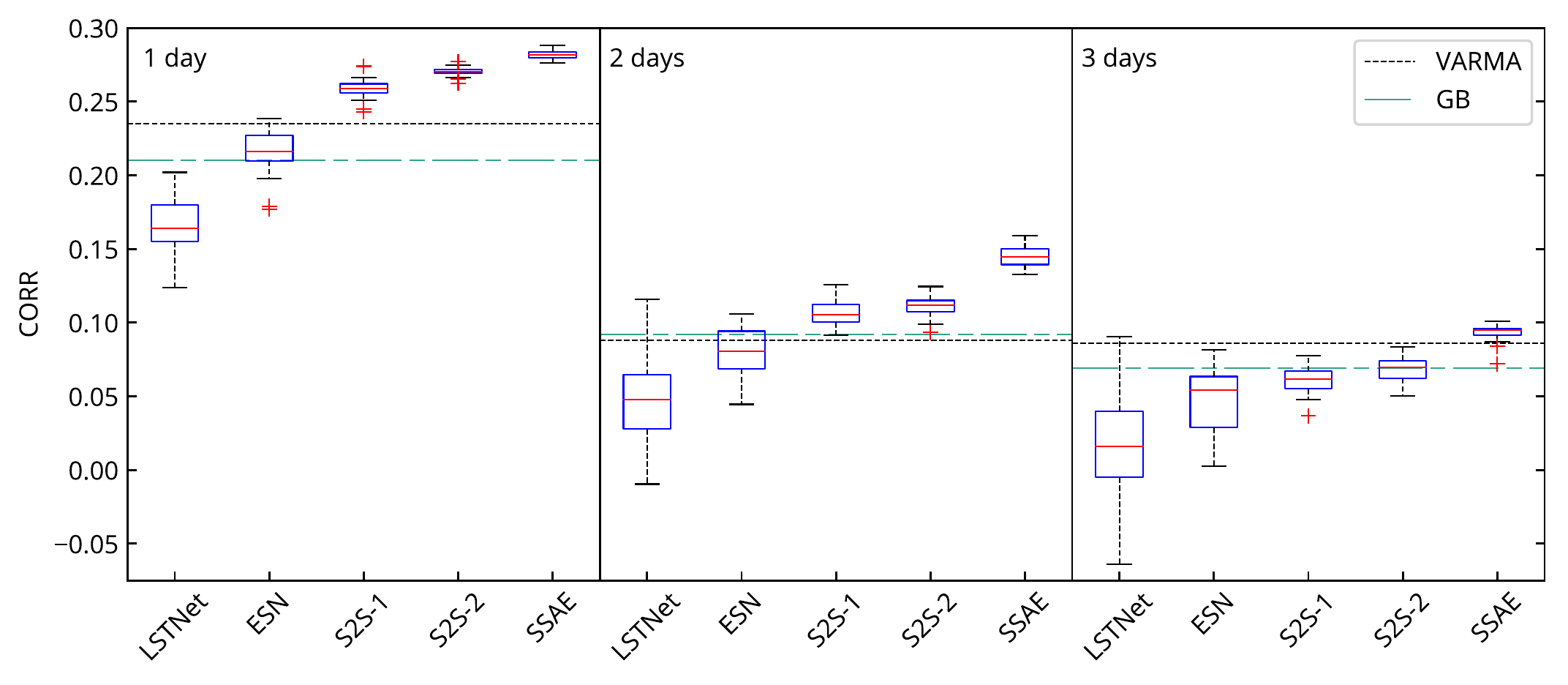}
    \caption{Performance of seven forecasting models on the test set. The scores of the SVR and TCN are not shown as they performed significantly worse than these models.}
    \label{fig:pvdbox}
\end{figure}

The RMSE and CORR scores on the test set are shown in \Cref{table:pvresults} and \Cref{fig:pvdbox}. Again, the SSAE outperforms other models in both metrics. Specifically, its RMSEs are $0.2\%-0.4\%$ smaller than those of S2S--2. One key observation entailed by the results is the significant improvement in CORR over those of the S2S--2, which is $4\%$ at horizon 1, $29\%$ at horizon 2 and $37\%$ at horizon $3$. As was the case with the previous experiment, the output variances of the SSAE and S2S--2 are smaller than those of the other deep learning models.

 \begin{table}[!htp]
    \centering\small
    \begin{tabular}{@{}lllllllll@{}}
       \toprule 
        & \multicolumn{2}{c}{Day 1}                           & \multicolumn{1}{c}{} & \multicolumn{2}{c}{Day 2}                           & \multicolumn{1}{c}{} & \multicolumn{2}{c}{Day 3}                           \\ \cline{2-3} \cline{5-6} \cline{8-9} 
Model   & \multicolumn{1}{c}{RMSE} & \multicolumn{1}{c}{CORR} & \multicolumn{1}{c}{} & \multicolumn{1}{c}{RMSE} & \multicolumn{1}{c}{CORR} & \multicolumn{1}{c}{} & \multicolumn{1}{c}{RMSE} & \multicolumn{1}{c}{CORR} \\ \midrule
VARMA     & 8.56 & 0.235 &    &  8.78 & 0.088 &  & \textbf{8.79} & 0.086  \\
SVR     & 9.23 & 0.138 &   & 9.42 & 0.059 &  & 9.42  & 0.055   \\
GB     & 8.62 & 0.210 &   & 8.81 & 0.092 &  & 8.82 & 0.069    \\
TCN     &    8.98\sg{0.33} & 0.141\sg{0.048} & & 8.85\sg{0.03} & 0.002\sg{0.040} & & 8.87\sg{0.04} & -0.027\sg{0.036}          \\
LSTNet     & 8.85\sg{0.07} & 0.166\sg{0.019} & & 9.00\sg{0.16} & 0.046\sg{0.029} & & 9.10\sg{0.32} & 0.017\sg{0.036}      \\
ESN     & 8.62\sg{0.04} & 0.216\sg{0.015} & & 8.85\sg{0.04} & 0.080\sg{0.017} & & 8.97\sg{0.11} & 0.048\sg{0.022}    \\
S2S--1  & 8.51\sg{0.01} & 0.259\sg{0.006} & & 8.77\sg{0.01} & 0.107\sg{0.009} & & 8.82\sg{0.01} & 0.061\sg{0.009}     \\
S2S--2  & 8.49\sg{0.01} & 0.270\sg{0.003} & & 8.77\sg{0.01} & 0.112\sg{0.007} & & 8.81\sg{0.01} & 0.068\sg{0.008}    \\
\textbf{SSAE} &  \textbf{8.46}\sg{0.01} & \textbf{0.282}\sg{0.003} & & \textbf{8.73}\sg{0.01} & \textbf{0.145}\sg{0.006} & & \textbf{8.79}\sg{0.01} & \textbf{0.093}\sg{0.005}     \\ 
\bottomrule
\end{tabular}%
\caption{The RMSE and CORR scores of day 1--3 forecasts. For all deep learning models, the sample mean and standard deviation of the scores from 30 training sessions are reported.}
\label{table:pvresults}
\end{table}

\begin{figure}[!t]
    \centering
    \includegraphics[width=\columnwidth]{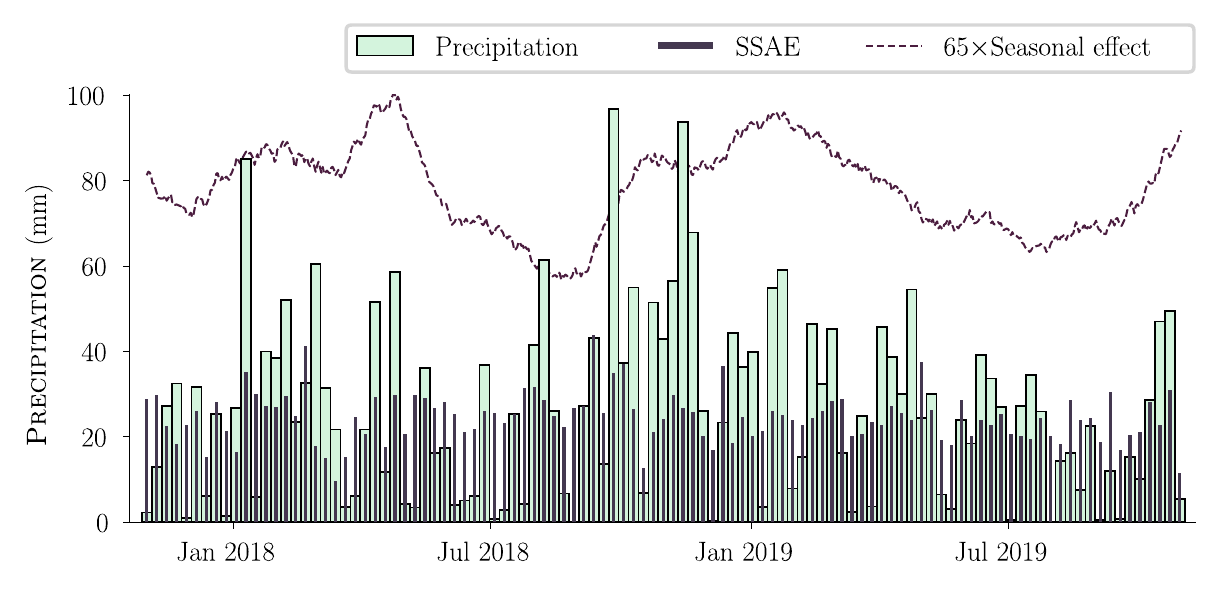}
    \caption{Weekly aggregated precipitation (mm), SSAE's next-day forecasts and the outputs of seasonal autoencoder from November 1st 2017 to October 31st 2019.}
    \label{fig:pvpred}
\end{figure}

\subsubsection*{Variable importance}

\begin{figure}[!t]
    \centering
    \includegraphics[width=\columnwidth]{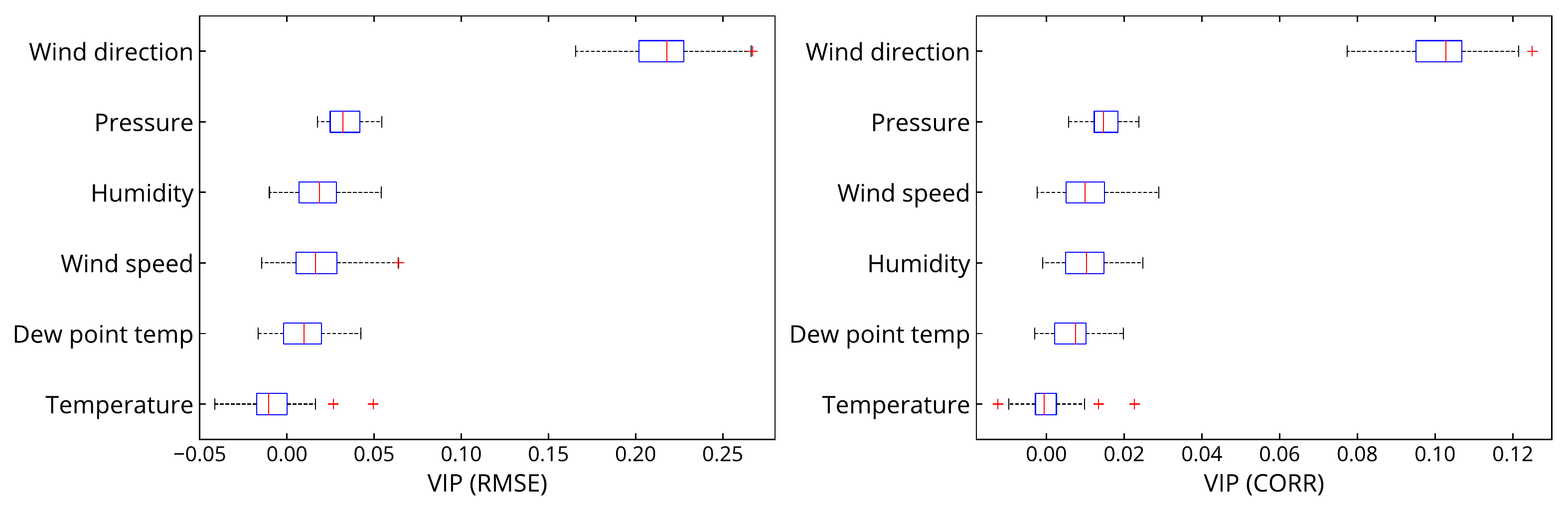}
    \caption{The variable importance scores measured by the RMSE and CORR scores of next-day forecasts on the test set.}
    \label{fig:providence_vip}
\end{figure}

The order of variable importance here might be different than that of the previous experiment. Let $R_j$ and $C_j$ be the new RMSE and CORR obtained after shuffling the variable $X_j$. The importance score of $X_j$ with respect to RMSE is  
\begin{align*}
    \text{VIP}^{\text{R}}_j &= R_j-8.46,
    \intertext{and the importance score of $X_j$ with respect to CORR is}
    \text{VIP}^{\text{C}}_j &= 0.282-C_j.
\end{align*}
The results of 30 repeated training sessions, shown in \Cref{fig:providence_vip}, indicate that the wind direction is the most important variable.

\subsubsection*{Predictive uncertainty}

\begin{figure}[!t]
    \centering
    \includegraphics[width=0.9\columnwidth]{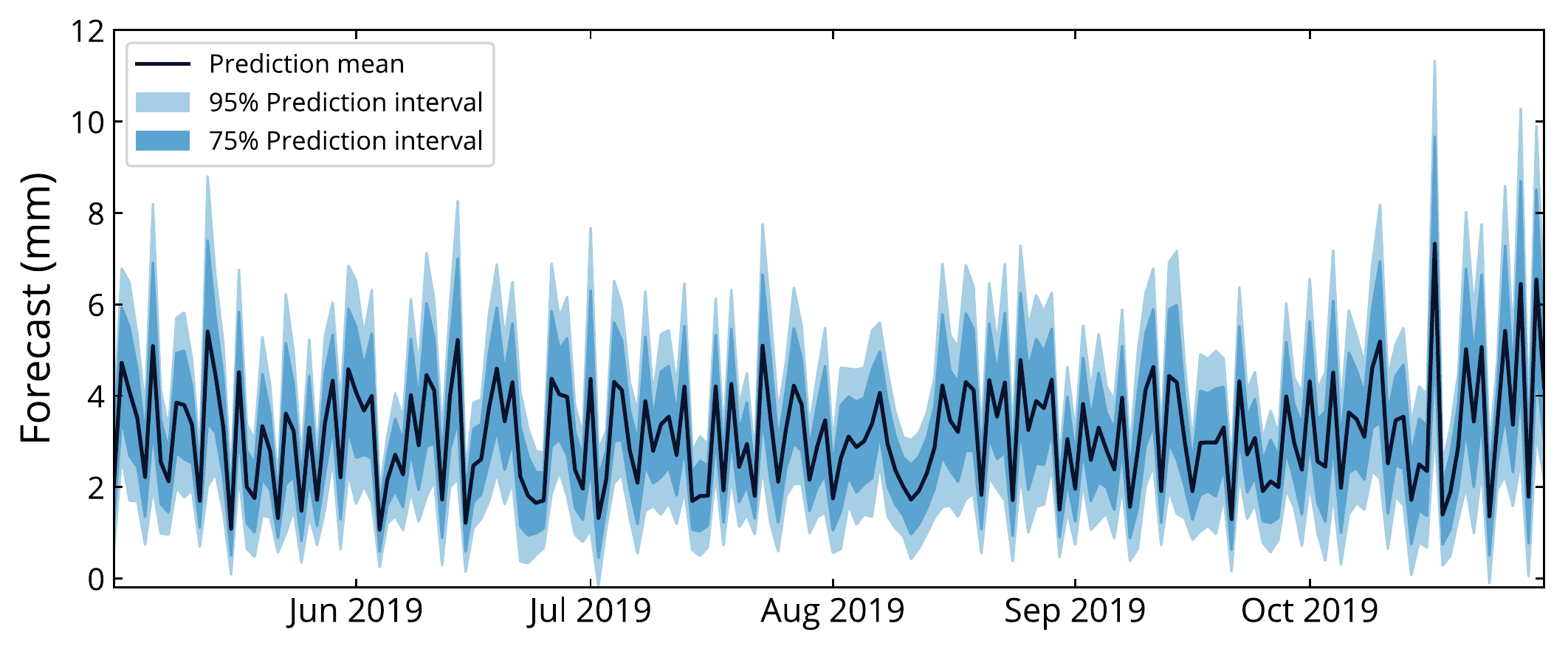}
    \caption{The confidence intervals of the next-day forecasts obtained via Monte Carlo dropout.}
    \label{fig:providence_conf}
\end{figure}

As in the previous experiment, we adopt the Monte Carlo dropout for measuring the predictive uncertainty of SSAE. We ran SSAE 50 times on the test set with $p=0.25$. The $75\%$ and $95\%$ confidence intervals of the forecasts are shown in \Cref{fig:providence_conf}.

\section{Discussion}

In both experiments, the SSAE outperforms all models in both metrics at horizons 1, 2, and 3. Specifically, it has higher RMSE and CORR scores than the original LSTM autoencoder (S2S--1), which is the current state-of-the-art in many time series forecasting \citep{s2s01,Xiang2020,Kao2020}. Moreover, our modification of the LSTM autoencoder (S2S--2) also outperforms the S2S--1, suggesting that splitting the LSTM unit in the decoder and replacing ReLU by $\tanh$ helped improve the forecasts. We also observe that, at horizons 2 and 3, there is a significant improvement in the CORR over the S2S--2, indicating that the seasonal autoencoder dominates the short-term counterpart.

Comparing the results from both experiments, we notice that the SSAE performs better in the first experiment; this is because the first dataset has clear and pronounced seasonality. As shown in \Cref{fig:cmpred}, the precipitation in Chiang Mai has a high contrast between the dry season (from January to April and from November to December) and the rainy season (from May to October). For this reason, the short-term component of the SSAE was able to differentiate between these two seasons just by using the data from past $6$ days. The main functionality of the seasonal autoencoder was to help improve forecasting non-zero amounts of precipitation during the dry season. For example, the seasonal autoencoder was able to improve the prediction of the single spike at the beginning of January 2018. As a consequence, the seasonal effect is high during the dry season and low during the wet season.
 
On the other hand, with the lack of dry season, the amount of precipitation in Providence is evenly distributed throughout the year. Without any short-term pattern, there is no benefit for the short-term component to look back further than two days. Consequently, the model had to rely on the seasonal autoencoder to learn the trend of the data. We can see from \Cref{fig:pvpred} that, in contrast to the first experiment, the seasonal effect matches the overall trend of the precipitation.

Going back to the first experiment, it is not surprising that, in the tropical climate, the humidity would be a suitable input variable for the seasonal autoencoder and the most important variable of the SSAE (\Cref{fig:weather_vip}). In the second experiment, the wind direction is the best variable for the seasonal autoencoder and the most important variable of the SSAE (\Cref{fig:providence_vip}). This result agrees with the studies by \cite{wind03,wind02,wind01} that the wind direction has significant effects on the precipitation, especially in the coastal and mountainous areas.

We also note the output variances in all deep learning models due to the random weight initialization (\Cref{table:cmresults} and \Cref{table:pvresults}). Both experiments demonstrate that the output variances of the S2S--1, S2S--2, and SSAE are significantly smaller than those of the other deep learning models. For TCN and LSTNet, this is not surprising as they have more parameters than the autoencoders. Even though ESN has fewer parameters than the autoencoders, its random weights in the hidden layer are fixed during the training stage, leading to a higher value of output variance compared to neural networks with similar configurations. Hence, not only does SSAE performs better, but it also has a smaller output variance compared to other deep learning models.

We now discuss the predictive uncertainties. Note that the forecasts in the second experiment (\Cref{fig:providence_conf}) fluctuate a lot more than those in the first experiment (\Cref{fig:weather_conf}). Despite this, the $95\%$ confidence intervals are mostly stable at $\pm$1 mm--2 mm around the means on both test sets, indicating that most of the predictive uncertainties are the parameter uncertainties. Nonetheless, the model is less confident in predicting a sudden change in precipitation, as evidenced by the $\pm$4mm-6mm intervals around the spikes.

\section{Future directions}

In both experiments, we can see from \Cref{fig:cmpred} and \Cref{fig:pvpred} that most forecast values of SSAE (and other models) underestimate the actual amount of precipitation. This is because the process of minimizing the MSE, combined with the sparsity nature of the data, encourages the model to give near-zero predictions. One way to put more emphasis on extreme values is to remove a fraction of instances with no precipitation, and the other is to replace MSE \eqref{eqn:mse} by a quantile loss function \citep{quantile} :
\begin{equation} q_r(\widehat{Y}_{N_i},Y_{N_i}) = r(Y_{N_i}-\widehat{Y}_{N_i})_{+}+(1-r)(\widehat{Y}_{N_i}-Y_{N_i})_{+},\label{quantile}\end{equation} 
where $(\cdot)_{+}=\max(0,\cdot)$ and $0\leq r\leq 1$. For $r$ close to one, the loss function penalizes the model for underestimating the precipitation. We have experimented on both datasets and found that training SSAE with the quantile loss helps increase the range of the forecasts and improve CORR in some cases (see \ref{sec:quantile}). We believe that this method could be useful for any task that requires predicting extreme values such as flood forecasting. 

In this study, we incorporated the seasonality into the deep learning model via a direct multiplication. To extend this idea, one might try different combinations between the outputs $g_{L}$ and $g_{S}$ of the seasonal and short-term network, respectively. For instance, we have experimented with one of the following models (see \ref{sec:combination}):
\begin{itemize}
      \item Additive: $Y_t = g_{L}(X_t)+g_{S}(X_t)$
    \item Linear combination: $Y_t=ag_{L}(X_t)+bg_{S}(X_t)$ where $a,b$ are trainable parameters,
\end{itemize}
and it turned out that the performances of these models were marginally worse than the multiplicative method. To come up with a combination of $g_L$ and $g_S$ that significantly improves upon these methods, one might need to rely on expert knowledge in weather modeling and time-series forecasting.

\section{Conclusions}
\label{sec:conclusions}
In this study, we propose a seasonally-integrated autoencoder (SSAE) for daily precipitation forecasting. The model consists of two long short-term memory (LSTM) autoencoders: one for learning short-term dynamics, and one for capturing the long-term seasonality. Moreover, we have modified the LSTM autoencoder to improve forecasting accuracy. Our experimental results show that, regardless of the climate type, our modified version of the LSTM autoencoder outperforms various classical and deep learning models. Secondly, by introducing the seasonal autoencoder to the model, we can further improve the correlation between the forecast and the actual values, especially when there is no clear wet and dry season. Another advantage of the SSAE is its small number of parameters compared to other deep learning models, which results in a smaller variance of outputs. From variable importance analysis, we find that the humidity and wind direction is the most important input variables of the SSAE for precipitation forecasting.

Besides the daily precipitation, there are many hourly and daily data that exhibit short-term nonlinear dynamic and long-term recurring patterns. For any future studies that aim to develop forecasting methods on such data, we believe that this study would provide not only with a potentially accurate model, but also with an insight on how to integrate seasonality into the model.

\section*{Funding}
This research is financially supported by Chiang Mai university. 

\section*{Acknowledgements}
The author would like to thank Thai Meteorological Department (\url{https://www.tmd.go.th}) and the National Oceanic and Atmospheric Administration (\url{https://www.ncdc.noaa.gov}) for providing the data.

\section*{Data Availability}
Datasets related to this article can be found at \url{http://dx.doi.org/10.17632/95mr7pr8rj.8}, an open-source online data repository hosted at Mendeley Data \citepalias{donlapark20}.

\section*{\refname}

\bibliographystyle{elsarticle-num}
\bibliography{Ponnoprat2020ASOC}








\appendix

\section{Training algorithm}\label{sec:trainalg}
In addition to minibatch training, there are several deep learning techniques that we employed to stabilize model training.
\subsection{Learning rate schedule}
The training of a deep learning model is usually done in multiple epochs where the data is shuffled and resplit after each epoch, introducing a stochastic element to the training. Because of this, we have to worry about the weights moving in a wrong direction from a bad batch sample. Hence, the concept of adaptive learning rate was thus introduced. Among the most popular ones is \emph{Adam} \citep{kingma14}, which suggests incorporating estimates of first and second moment of the (stochastic) gradients, say $\hat{r}_{w,k}$ and $\hat{s}_{w,k}$ respectively, in the weight update \eqref{update} as follows:
\begin{equation}\label{eta}
    w_{k} = w_{k-1}-\frac{\eta}{\sqrt{\hat{s}_{w,k}}}\hat{r}_{w,k}.  
\end{equation}
Hence, the new learning rate $\eta_{w,k}=\eta/\sqrt{\hat{s}_{w,k}}$ of the weights with large partial derivatives is smaller than that with lower partial derivatives, resulting in the more gently sloped of the weights movements. 

However, the result of model training can still be quite sensitive to the choice of the initial learning rate $\eta$. To reduce the search time for an optimal $\eta$, we opt for \emph{RAdam}, a modification of Adam that is less sensitive to the choice of learning rate $\eta$, in all of our experiments. For technical details of the implementation of RAdam, see \cite{liu19}.

\subsection{Exponential learning rate decay}
An additional technique that we used to make sure that the weights escape from the local minima is to set the initial learning rate $\eta_{w,k}$ to be high and continuously decrease it until the weights are stabilized. This can be done by imposing an exponential decay on the learning rate.
\begin{equation}\label{eta1}
    w_{k} = w_{k-1}-\frac{\eta\epsilon^k}{\sqrt{\hat{s}_{w,k}}}\hat{r}_{w,k}.  
\end{equation}
where $\epsilon$ is the decay rate. We can find the right number of epochs after which the weights are stabilized by looking at the MSE of the validation set.

\subsection{Weight initialization}
To make sure that the size of the output does not grow very large or completely vanish through multiple layers, an appropriate weight initialization ($w_0$) is required. Moreover, since the loss function of deep learning is typically non-convex, we need good control over learning rate schedule to ensure that the weights are not stuck at local minima. Regarding the weight initialization, we employed \emph{Glorot normal} (also known as \emph{Xavier normal}) scheme. Suppose that $n_j$ and $n_{j+1}$ are the number of nodes in the $j$-th and $j+1$-th layer, respective. Then the weights connecting these two layers are randomly drawn from the truncated normal distribution (i.e. any value more than two standard deviation from the mean are discarded and redrawn)
\[w\sim N\left(0,\frac{2}{n_j+n_{j+1}}\right),\]
in order to maintain the size of the gradient in \eqref{eta1}, and hence the size of the weights and the outputs. 

In the training of SSAE on Chiang Mai dataset, the initial learning rate of $\eta$ was $1\times 10^{-3}$ and the decay rate $\epsilon$ was $0.955/30$. On the Providense dataset, the initial learning rate was $8\times 10^{-4}$ and the decay rate was $0.955/20$. 

\section{Hyperparameter settings}
\label{sec:hyper}
The hyperparameter settings for all models in our experiments besides SSAE are shown in \Cref{table:hyper1} and \Cref{table:hyper2}. We observe that the length of the look-back window $T$ varies across the models. The hyperparameters for VARMA were found via a grid search, while those for the rest of the models were found via random searches.

\begin{table}[!htbp]
\centering
\begin{tabular}{@{}lll@{}}
\toprule
Model & $T$ & Other parameters \\ \hline
VARMA & 2 & \begin{tabular}[c]{@{}l@{}}$\text{AR} = 2$, $\text{MA} = 1$\end{tabular} \\ \hline
SVR & 40 & \begin{tabular}[c]{@{}l@{}}$C=9$, $\epsilon=0.053$\end{tabular} \\ \hline
GB & 65 & \begin{tabular}[c]{@{}l@{}}$\text{learning rate}=0.0565$\\ $\text{maximum depth}=3$\\ $\text{number of trees}=130$\\ $L_2 \text{ regularization weight}=35$\end{tabular} \\ \hline
TCN & 64 & \begin{tabular}[c]{@{}l@{}}$\text{number of filters}=4$\\ $\text{convolutional kernel size}=3$\\ $\text{dilations}=(1,2,4,8,16)$\\ $\text{number of stacks}=3$\end{tabular} \\ \hline
LSTNet & 16 & \begin{tabular}[c]{@{}l@{}}$\text{number of filters}=84$\\ $\text{convolution kernel size}=3$\\ $\text{dimension of GRU's hidden states}=106$\\ $\text{dimension of SkipGRU's hidden states}=12$\\ $\text{number of time steps to skip}=8$\\ $\text{look-back period for the autoregression} = 4$\end{tabular} \\ \hline
ESN & 300 & \begin{tabular}[c]{@{}l@{}}$\text{dimension of the hidden layer}=100$\\ $\text{reservoir's spectral radius}=0.7776$\\ $\text{leaking rate}=1.1668$\end{tabular} \\ \hline
S2S--1 & 6 & $\text{number of hidden units}=100$ \\ \hline
S2S--2 & 6 & $\text{number of hidden units}=100$ \\ \bottomrule
\end{tabular}%
\caption{The settings for the competing models in the first experiment.}
\label{table:hyper1}
\end{table}

\begin{table}[!htbp]
    \centering
    \begin{tabular}{@{}lll@{}}
\toprule
Model & $T$ & Other parameters \\ \hline
VARMA & 2 & \begin{tabular}[c]{@{}l@{}}$\text{AR} = 2$, $\text{MA} = 0$\end{tabular} \\ \hline
SVR & 140 & \begin{tabular}[c]{@{}l@{}}$C=3.62$, $\epsilon=0.077$\end{tabular} \\ \hline
GB & 40 & \begin{tabular}[c]{@{}l@{}}$\text{maximum depth}=2$\\ $\text{learning rate}=0.058$\\ $\text{number of trees}=120$\\ $L_2 \text{ regularization weight}=20$\end{tabular} \\ \hline
TCN & 8 & \begin{tabular}[c]{@{}l@{}}$\text{number of filters}=4$\\ $\text{convolutional kernel size}=5$\\ $\text{dilations}=(1,2,4)$\\ $\text{number of stacks}=1$\end{tabular} \\ \hline
LSTNet & 63 & \begin{tabular}[c]{@{}l@{}}$\text{number of filters}=65$\\ $\text{convolution kernel size}=3$\\ $\text{dimension of GRU's hidden states}=84$\\ $\text{dimension of SkipGRU's hidden states}=12$\\ $\text{number of time steps to skip}=30$\\ $\text{look-back period for the autoregression} = 1$\end{tabular} \\ \hline
ESN & 250 & \begin{tabular}[c]{@{}l@{}}$\text{dimension of the hidden layer}=56$\\ $\text{reservoir's spectral radius}=0.03336$\\ $\text{leaking rate}=1.5283$\end{tabular} \\ \hline
S2S--1 & 2 & $\text{number of hidden units}=100$ \\ \hline
S2S--2 & 2 & $\text{number of hidden units}=120$ \\ \bottomrule
\end{tabular}
\caption{The settings for the competing models in the second experiment.}
\label{table:hyper2}
\end{table}

\section{Comparison between MSE and quantile loss functions}
\label{sec:quantile}

As an alternative to MSE, we study the effect of training SSAE with the quantile loss function \citep{quantile}.
\begin{equation} q_r(\widehat{Y}_{N_i},Y_{N_i}) = r(Y_{N_i}-\widehat{Y}_{N_i})_{+}+(1-r)(\widehat{Y}_{N_i}-Y_{N_i})_{+},\label{quantile}\end{equation} 
where $(\cdot)_{+}=\max(0,\cdot)$ and $0\leq r\leq 1$. For $r$ close to one, the loss function penalizes the model for underpredicting the precipitation. 

\begin{figure}[!htbp]
    \centering
    \includegraphics[width=\columnwidth]{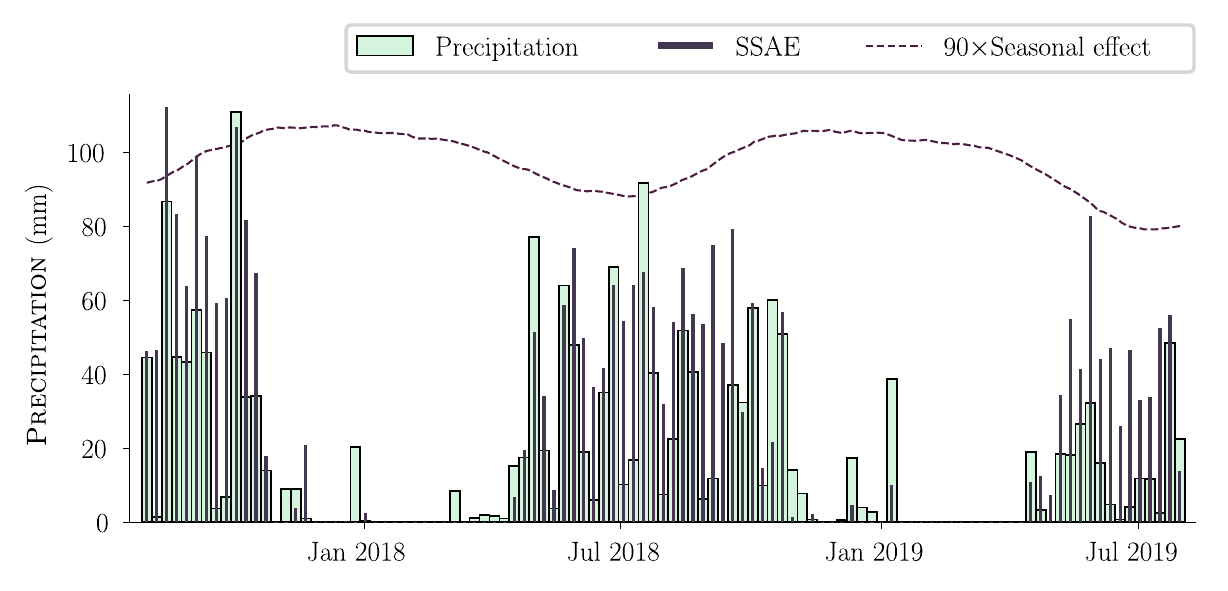}
    \caption{The plot of aggregated precipitation forecast at Chiang Mai International Airport from August 1st 2015 to July 31st 2019. The model was trained using $q_{0.8}$ as the loss function.}
    \label{fig:cm_q80}
\end{figure}

\begin{figure}[!htbp]
    \centering
    \includegraphics[width=\columnwidth]{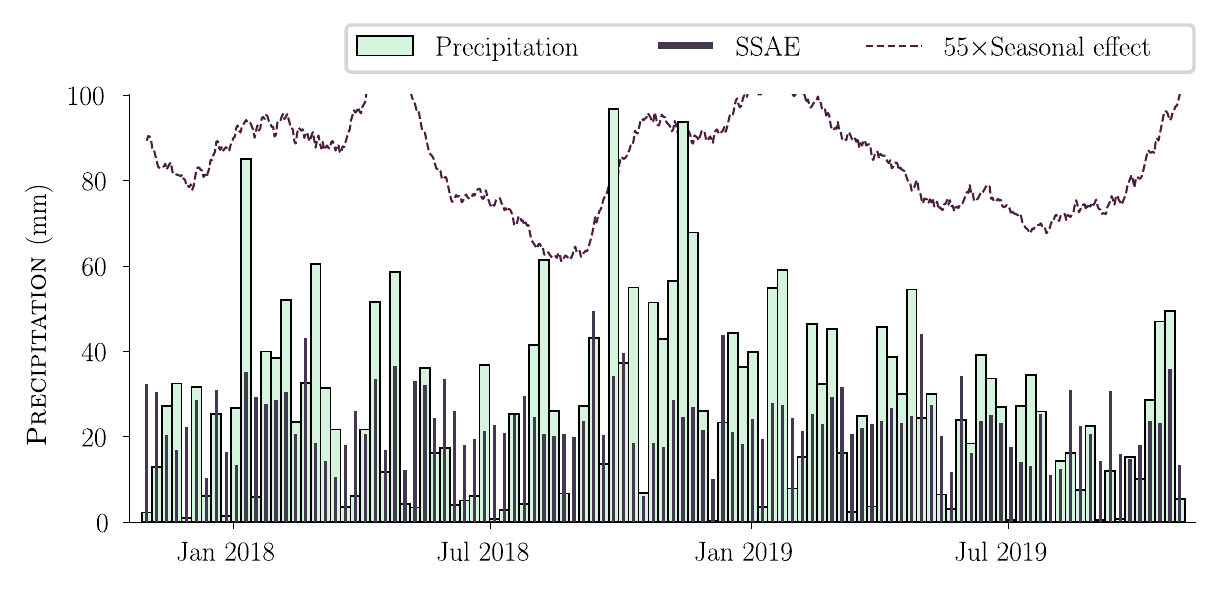}
    \caption{The plot of aggregated precipitation forecast at Theodore Francis Green State Airport from November 1st 2015 to October 31st 2019. The model was trained using $q_{0.8}$ as the loss function.}
    \label{fig:pvd_q80}
\end{figure}

In this experiment, we trained SSAE for day-1 forecast on both datasets with different values of $r$. The results are shown in \Cref{table:quantile}. For both datasets, the best value of $r$ is $0.8$, and there is an improvement in CORR over the model trained with MSE on the Providence dataset. The forecast values on Chiang Mai dataset (\Cref{fig:cm_q80}) are overall higher compared to SSAE trained with MSE (\Cref{fig:cmpred}). In contrast, the forecasts of both models on the Providence dataset stay roughly on the same level (\Cref{fig:pvd_q80} and \Cref{fig:pvpred}). The author conjectures that this is because the weather variables in the Providence dataset are not sufficient to explain the variance in the precipitation. 

\begin{table}[htbp]
    \centering
    \begin{tabular}{@{}llllll@{}}
       \toprule 
        & \multicolumn{2}{c}{Chiang Mai}                           & \multicolumn{1}{c}{} & \multicolumn{2}{c}{Providence}                                                     \\ \cline{2-3} \cline{5-6}  
Loss function   & \multicolumn{1}{c}{RMSE} & \multicolumn{1}{c}{CORR} & \multicolumn{1}{c}{} & \multicolumn{1}{c}{RMSE} & \multicolumn{1}{c}{CORR}  \\ \midrule
MSE &  \textbf{6.96}\sg{0.01} & \textbf{0.418}\sg{0.002} & & \textbf{8.46}\sg{0.01} & 0.282\sg{0.003}     \\ 
$q_{0.6}$     & 7.35\sg{0.01} & 0.391\sg{0.004} & & 9.11\sg{0.03} & 0.249\sg{0.009}    \\
$q_{0.7}$     & 7.02\sg{0.02} & 0.411\sg{0.005} & & 11.68\sg{0.26} & 0.286\sg{0.002}  \\
$q_{0.8}$  & 7.40\sg{0.04} & 0.417\sg{0.003} & & 8.61\sg{0.07} & \textbf{0.292}\sg{0.002}  \\
$q_{0.9}$  & 11.26\sg{0.15} & 0.398\sg{0.005} & & 11.67\sg{0.20} & 0.287\sg{0.003}  \\
\bottomrule
\end{tabular}%
\caption{The RMSE and CORR scores of the next-day forecast from SSAE trained with different loss functions. For each loss function, the mean and sample standard deviation of the scores from 30 training sessions are reported.}

\label{table:quantile}
\end{table}

\section{Combining the seasonal and short-term outputs in SSAE}
\label{sec:combination}
We compare the performance of SSAE with two additional autoencoder models that have the same seasonal and short-term components ($g_L$ and $g_S$) combined in different ways: 
\begin{itemize}
    \item SSAE$_+$:  $Y_t = g_{L}(X_t)+g_{S}(X_t)$
    \item SSAE$_l$: $Y_t=ag_{L}(X_t)+bg_{S}(X_t)$, where $a,b$ are parameters that can be trained from the data.
\end{itemize}

We trained each of the models for 30 repetitions and recorded the average and sample standard deviation of RMSE and CORR scores. The results on both datasets are shown in \Cref{table:cmfunc} and \Cref{table:pvfunc}. We see that these models have similar performance, but compared to other models, SSAE is marginally better across all horizons in terms of these two scores.

\begin{table}[htbp]
    \centering
    \begin{tabular}{@{}lllllllll@{}}
       \toprule 
        & \multicolumn{2}{c}{Day 1}                           & \multicolumn{1}{c}{} & \multicolumn{2}{c}{Day 2}                           & \multicolumn{1}{c}{} & \multicolumn{2}{c}{Day 3}                           \\ \cline{2-3} \cline{5-6} \cline{8-9} 
       Model   & \multicolumn{1}{c}{RMSE} & \multicolumn{1}{c}{CORR} & \multicolumn{1}{c}{} & \multicolumn{1}{c}{RMSE} & \multicolumn{1}{c}{CORR} & \multicolumn{1}{c}{} & \multicolumn{1}{c}{RMSE} & \multicolumn{1}{c}{CORR} \\ \midrule
S2S--2  & 6.98\sg{0.01} & 0.410\sg{0.004} & & 7.30\sg{0.01} & 0.305\sg{0.003} & & 7.38\sg{0.01} & 0.276\sg{0.005}    \\
SSAE &  \textbf{6.96}\sg{0.01} & \textbf{0.418}\sg{0.002} & & \textbf{7.28}\sg{0.01} & \textbf{0.311}\sg{0.003} & & \textbf{7.35}\sg{0.01} & \textbf{0.287}\sg{0.006}     \\ 
SSAE$_+$  & 7.00\sg{0.01} & 0.409\sg{0.004} & & 7.30\sg{0.01} & 0.309\sg{0.004} & & 7.38\sg{0.02} & 0.279\sg{0.007}    \\
SSAE$_l$ &  6.98\sg{0.02} & 0.412\sg{0.005} & & 7.30\sg{0.01} & 0.308\sg{0.004} & & 7.38\sg{0.02} & 0.278\sg{0.005}     \\ 
\bottomrule
\end{tabular}%
\caption{The RMSE and CORR scores of day 1--3 forecasts on the Chiang Mai dataset. The mean and sample standard deviation of the scores from 30 training sessions are reported.}
\label{table:cmfunc}
\end{table}

\begin{table}[htbp]
    \centering
    \begin{tabular}{@{}lllllllll@{}}
       \toprule 
        & \multicolumn{2}{c}{Day 1}                           & \multicolumn{1}{c}{} & \multicolumn{2}{c}{Day 2}                           & \multicolumn{1}{c}{} & \multicolumn{2}{c}{Day 3}                           \\ \cline{2-3} \cline{5-6} \cline{8-9} 
Model   & \multicolumn{1}{c}{RMSE} & \multicolumn{1}{c}{CORR} & \multicolumn{1}{c}{} & \multicolumn{1}{c}{RMSE} & \multicolumn{1}{c}{CORR} & \multicolumn{1}{c}{} & \multicolumn{1}{c}{RMSE} & \multicolumn{1}{c}{CORR} \\ \midrule
S2S--2  & 8.49\sg{0.01} & 0.271\sg{0.003} & & 8.77\sg{0.01} & 0.113\sg{0.007} & & 8.81\sg{0.01} & 0.068\sg{0.007}    \\
SSAE &  \textbf{8.46}\sg{0.01} & \textbf{0.282}\sg{0.003} & & \textbf{8.73}\sg{0.01} & \textbf{0.145}\sg{0.006} & & \textbf{8.79}\sg{0.01} & \textbf{0.093}\sg{0.005}     \\ 
SSAE$_+$     & 8.33\sg{0.01} & 0.272\sg{0.003} & & 8.59\sg{0.01} & 0.130\sg{0.008} & & 8.64\sg{0.01} & 0.082\sg{0.005}    \\
SSAE$_l$  & 8.33\sg{0.01} & 0.272\sg{0.003} & & 8.60\sg{0.01} & 0.126\sg{0.008} & & 8.64\sg{0.01} & 0.079\sg{0.008}     \\
\bottomrule
\end{tabular}%
\caption{The RMSE and CORR scores of day 1--3 forecasts on the Providence dataset. The mean and sample standard deviation of the scores from 30 training sessions are reported.}
\label{table:pvfunc}
\end{table}

\end{document}